\documentclass[10pt,journal,compsoc]{IEEETran}

\usepackage{subfiles}

\ifCLASSOPTIONcompsoc
  \usepackage[nocompress]{cite}
\else
  \usepackage{cite}
\fi
\usepackage{graphicx}

\usepackage{tikz}
\usepackage{comment}
\usepackage{amsmath,amssymb} 
\usepackage{color}
\usepackage{booktabs}

\usepackage[accsupp]{axessibility}  

\usepackage{bbm}
\usepackage{hyperref}
\usepackage{amsfonts}
\usepackage{multirow}
\usepackage{pbox}
\usepackage[normalem]{ulem}
\usepackage{ulem}

\usepackage{xcolor}
\usepackage{enumerate}
\usepackage{diagbox}
\usepackage{bm}
\usepackage{enumitem}

\newcommand{\argmin}{\mathop{\mathrm{argmin}}}

\newcommand{\norm}[1]{\left\|#1\right\|} 
\newcommand{\midsepremove}{\aboverulesep = 0mm \belowrulesep = 0mm}
\newcommand{\midsepdefault}{\aboverulesep = 0.605mm \belowrulesep = 0.984mm}

\begin{document}

\title{Contrastive Learning for Joint Normal Estimation\\and Point Cloud Filtering}

\author{Dasith~de~Silva~Edirimuni,~
        Xuequan~Lu,~\IEEEmembership{Senior~Member,~IEEE,}
        Gang~Li,~\IEEEmembership{Senior~Member,~IEEE,}
        and~Antonio~Robles-Kelly,~\IEEEmembership{Senior~Member,~IEEE}
\IEEEcompsocitemizethanks{\IEEEcompsocthanksitem D. de Silva Edirimuni, X. Lu, G. Li and A. Robles-Kelly are with the School of Information Technology, Deakin University, Waurn Ponds, Victoria, 3216, Australia (e-mail: \{dtdesilva, xuequan.lu, gang.li, antonio.robles-kelly\}@deakin.edu.au). A. Robles-Kelly is also with the Defense Science and Technology Group, Australia.}
\thanks{Manuscript received Month Day, Year; revised Month Day, Year.}
\thanks{(Corresponding author: Xuequan Lu.)}}

\markboth{Journal of \LaTeX\ Class Files,~Vol.~X, No.~X, Month~Year}%
{de Silva Edirimuni \MakeLowercase{\textit{et al.}}: Contrastive Learning for Joint Normal Estimation and Point Cloud Filtering}

\IEEEtitleabstractindextext{%
\begin{abstract}
Point cloud filtering and normal estimation are two fundamental research problems in the 3D field. 
Existing methods usually perform normal estimation and filtering separately and often show sensitivity to noise and/or inability to preserve sharp geometric features such as corners and edges. In this paper, we  propose a novel deep learning method to jointly estimate normals and filter point clouds. 
We first introduce a 3D patch based contrastive learning framework, with noise corruption as an augmentation, to train a feature encoder capable of generating faithful representations of point cloud patches while remaining robust to noise.  
These representations are consumed by a simple regression network and supervised by a novel joint loss, simultaneously estimating point normals and displacements that are used to filter the patch centers. 
Experimental results show that our method well supports the two tasks simultaneously and preserves sharp features and fine details. It generally outperforms state-of-the-art techniques on both tasks. Our source code is available at \url{https://github.com/ddsediri/CLJNEPCF}.
\end{abstract}

}

\maketitle

\IEEEdisplaynontitleabstractindextext

%
\IEEEpeerreviewmaketitle

\IEEEraisesectionheading{\section{Introduction}\label{sec:introduction}}

%
%
%
%
\IEEEPARstart{P}{oint} clouds have numerous applications as they provide a natural representation of 3D geometric information. They have seen applications in fields such as autonomous driving, robotics, 3D printing and urban planning~\cite{Luo-Pillar-Motion,Bekiroglu-PCD-Robotics,Kim-3D-Printing,Urech-Urban-Planning}. Captured using 3D sensors, point clouds consist of unordered points which lack connectivity information between individual points. The captured point cloud information may be corrupted with noise. Therefore, one fundamental research problem is point cloud filtering, also known as denoising. Another fundamental task is normal estimation at individual points.  Together, they facilitate other tasks such as 3D rendering and surface reconstruction.


Conventional normal estimation methods, such as Principal Component Analysis (PCA) and its variants~\cite{Hoppe-PCA,Mitra-Estimating-Surface-Normals,Pauly-Point-Sampled,Yoon-Surface-Normal-Ensembles} and Voronoi diagram based approaches \cite{Amenta-Surface-Voronoi,Alliez-Voronoi-Variational,Dey-Voronoi-based-Normal-Estimation}, perform poorly when estimating the normals at sharp features such as corners or edges and show high sensitivity to noise. To address these issues, a number of learning based methods have been recently proposed such as Deep Feature Preserving (DFP) \cite{Lu-Deep-Feature-Preserving} and Nesti-Net~\cite{Ben-Shabat-Nesti-Net}. However, they have large network sizes and therefore are typically slow. Methods such as AdaFit~\cite{Zhu-AdaFit} and Deep Iterative (DI)~\cite{Lenssen-Deep-Iterative} offer more lightweight solutions that perform admirably, but still show less robust results at higher noise levels. 

Point cloud filtering can be classified into two main types: normal based methods~\cite{RIMLS-Oztireli,Sun-L0,Avron-L1,Lu-Deep-Feature-Preserving} and position based methods~\cite{Lipman-LOP,Huang-WLOP,Rakotosaona-PCN,Zhang-Pointfilter}. The former utilizes normal information at a given point in order to apply a position update  algorithm~\cite{Lu-Deep-Feature-Preserving}, while the latter does not require normal information and relies solely on position information. Among position based methods, a common issue is the inability to preserve sharp features during the filtering process while normal based methods rely heavily on normal accuracy. Learning based approaches seek to resolve this. In particular,  Pointfilter~\cite{Zhang-Pointfilter}, performs effectively at preserving sharp feature information on CAD-like shapes yet fails to generalize to large scenes. Methods such as PointCleanNet~\cite{Rakotosaona-PCN} and TotalDenoising~\cite{Hermosilla-Total-Denoising} also perform sub-optimally, tending to smear sharp features. 

In this paper, we propose a novel method capable of simultaneously inferring point normals and displacements while maintaining robustness to noise. Our method comprises of a feature encoder capable of generating latent representations of patches based on patch similarity and a regressor capable of inferring point normals and displacements simultaneously. We introduce a 3D patch based contrastive learning framework to train the feature encoder which employs noise corruption as an augmentation technique, allowing the encoder to identify the sharp geometric features of the underlying clean patch despite different levels of noise corruptions. The regressor consumes the latent representation of a patch and outputs the point normal and the displacement required to filter the central point of that patch. To train the regressor, we introduce a novel loss function that jointly penalizes inferred point position error and normal estimation error by exploiting the relationship between a point's position and normal. We intuitively assume that a filtered point's normal should correspond to a ground truth point's normal if this ground truth point first corresponds to that filtered point in position, thus leading to the relationship between filtering and normal estimation.

The main contributions of this paper are as follows. 
\begin{itemize}
    \item We develop a novel framework capable of inferring both points' displacements and normals simultaneously by introducing a loss function capable of constraining both filtering and normal estimation tasks. This joint loss penalizes both position regression error and normal estimation prediction error and allows the network to learn both filtered displacements and point normals.
    \item We introduce 3D patch based contrastive learning to generate effective patch-wise representations.
\end{itemize}
We conduct extensive experiments and demonstrate that our method, in general, outperforms state-of-the-art normal estimation and filtering techniques.

\section{Related work}

\textbf{Normal estimation.} In its earliest incarnation, normal estimation was based on Principal Component Analysis  (PCA)~\cite{Hoppe-PCA}. Several variants of this initial PCA method have also been proposed~\cite{Mitra-Estimating-Surface-Normals,Pauly-Point-Sampled,Yoon-Surface-Normal-Ensembles}. Thereafter, approaches based on Voronoi cells were used to reconstruct surfaces while preserving sharp features and estimating normals~\cite{Amenta-Surface-Voronoi,Alliez-Voronoi-Variational,Dey-Voronoi-based-Normal-Estimation}. Recently, Lu et al.~\cite{Lu-Low-Rank} proposed a normal estimation method based on a Low Rank Matrix Approximation (LRMA) algorithm. Additionally, methods such  as~\cite{Li-Robust-Sharp-Features,Zhang-Guided-Least-Squares,Zhang-Low-Rank,Zhang-Pair-Consistency-Voting} utilized point statistics and clustering to determine point normals.

\textbf{Normal estimation (learning-based)}. 
One of the first learning models for normal estimation, HoughCNN, employs a voting mechanism for estimating normals. They utilize a local patch representation in Hough space that can be consumed by a CNN~\cite{Boulch-HoughCNN}. However, with the advent of PointNet~\cite{Qi-PointNet} and PointNet++~\cite{Qi-PointNet++}, newer methods have been proposed that directly consume point sets. PCPNet is one such example, which consumes point cloud patches at multiple scales~\cite{Guerrero-PCPNet}. Similarly, Nesti-Net consumes patches at multiple scales but also employs multiple sub-networks, Mixture-of-Experts, that specialize in estimating normals at these scales~\cite{Ben-Shabat-Nesti-Net}. Wang and Prisacariu introduced NINormal, a self-attention based normal estimation scheme~\cite{Wang-NINormal} while Lu et al. proposed Deep Feature Preserving (DFP), a two step mechanism that classifies points into feature and non-feature points and, subsequently, estimates their normals based on this classification~\cite{Lu-Deep-Feature-Preserving}. Finally, several deep learning methods based on weighted least squares plane fitting have been proposed~\cite{Lenssen-Deep-Iterative,Ben-Shabat-DeepFit,Zhu-AdaFit}. While these methods focus on accurately determining unoriented normals, the work of Wang et al.~\cite{Wang-Deep-Global-NO} focuses on estimating point normals and their orientations.

\textbf{Point cloud filtering}. Traditional filtering applications center around Moving Least Squares (MLS) approaches~\cite{MLS-Levin,Kolluri-MLS}. Alexa et al.~\cite{Alexa--MLS-PSS} built on MLS techniques to minimize the approximation error of denoised point set surfaces. These methods perform poorly on point sets with sharp features, an issue that Adamson and Alexa~\cite{IMLS-Adamson} and Guennebaud and Gross~\cite{APSS-Guennebaud} aimed to tackle. Lipman et al. developed the Locally Optimal Projection (LOP) operator which does not depend on a local data parametrization such as a local normal or tangent plane~\cite{Lipman-LOP}. This projection operator was further enhanced by Huang et al. and Preiner et al., who proposed a Weighted LOP (WLOP)~\cite{Huang-WLOP} and Continuous LOP (CLOP)~\cite{Preiner-CLOP}, respectively. The main drawback to these MLS and LOP based techniques is their inability to preserve sharp features. Oztireli, Guennebaud and Gross~\cite{RIMLS-Oztireli} proposed Robust Implicit Moving Least Squares~\cite{RIMLS-Oztireli} which improves the filtering ability to preserve sharp features but relies heavily on the accuracy of normal information. Lu et al. proposed a point cloud filtering scheme based on normals estimated by their LRMA algorithm~\cite{Lu-Low-Rank}. Remil et al. reformulated point cloud filtering as a global, sparse optimization problem which is solved using Augmented Lagrangian Multipliers~\cite{Remil-Data-Driven-Sparse-Priors}.

\textbf{Point cloud filtering (learning-based)}. PointProNets used a CNN which consumes noisy height-maps and returns filtered ones~\cite{Roveri-PointProNets}. EC-Net employed a supervised scheme for edge aware filtering and upsampling~\cite{Yu-EC-Net}. PCN uses a $L_1$ norm loss based network to remove outliers and $L_2$ norm loss based network to filter points~\cite{Rakotosaona-PCN}. Pointfilter takes into account local structure by considering points and their ground-truth normals, during training time, to infer filtered positions~\cite{Zhang-Pointfilter}. DFP~\cite{Lu-Deep-Feature-Preserving} employs the position update mechanism of~\cite{Lu-Low-Rank} to filter points based on the estimated normals.  ScoreDenoise (SD) models a noisy point cloud's underlying surface with a 3D distribution supported by 2D manifolds and estimates the score for the gradient of the noise convolved distribution~\cite{Luo-Score-Based-Denoising}. TotalDenoising (TD) offers an unsupervised learning alternative to the aforementioned supervised schemes~\cite{Hermosilla-Total-Denoising}.

\textbf{Contrastive learning}. Recently, we have seen the increased use of contrastive learning in generating faithful representations based on similarity between inputs. Self-supervised learning that maximizes agreement between similar inputs was first proposed by Becker and Hinton~\cite{Becker-Contrastive}. Thereafter, contrastive learning was further exploited to learn lower dimensional representations of high dimensional image data by the work of Hadsell, Chopra and LeCun~\cite{Hadsell-Contrastive}. Chen et al. utilized more recent neural network architectures and data augmentation methods in their SimCLR method~\cite{Chen-SimCLR}. Although initially designed for 2D image processing tasks, contrastive learning is now seeing applications in 3D representation learning~\cite{jiang2021unsupervised,Lal-CoCoNets,Du-Self-Supervised-Point-Cloud,Afham-CrossPoint} and for specific point cloud processing tasks such as shape completion, segmentation and scene understanding~\cite{Alliegro-Shape-Completion,Li-HybridCR,Hou-3D-Scene}. However, it has never before been explored in terms of the problems of normal estimation and point cloud filtering, which we focus on in this work.

\begin{figure}[!ht]
\centering
\includegraphics[width=0.8\linewidth]{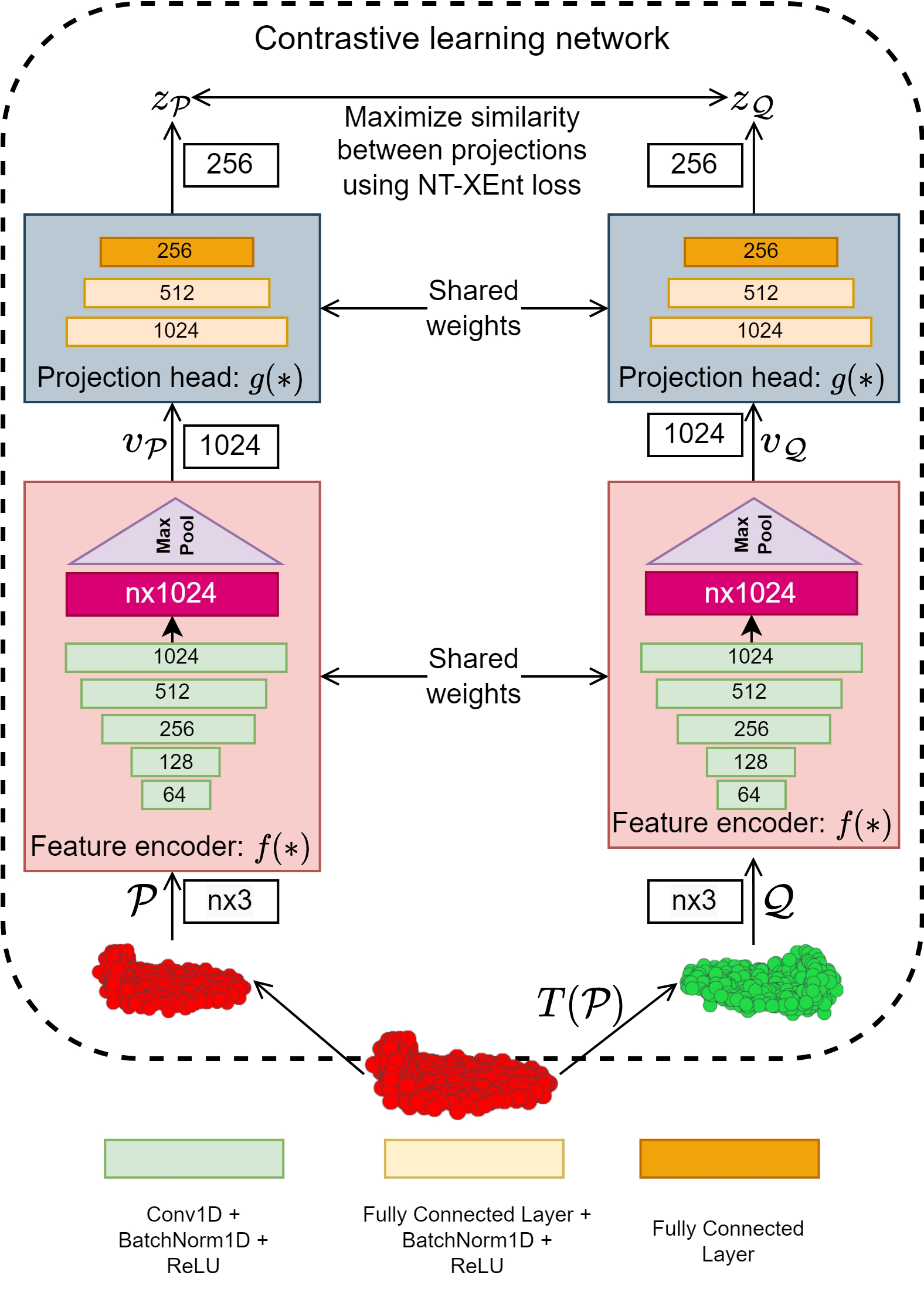}
\caption{
3D patch based contrastive learning. 
}
\label{fig:contrastive-learning-framework}
\end{figure}

\section{Background and Motivation}
In this section, we look at the motivation for our contrastive learning based joint normal estimation and filtering method.
\subsection{Patch-based contrastive learning}
\label{subsec:patch-based-cl}
As mentioned earlier, contrastive learning has emerged as an effective method of generating latent representations of inputs such as images or point clouds based on similarity between augmented pairs of inputs which are seen by the network during training~\cite{Chen-SimCLR}. The work of Xie et al.~\cite{Xie-PointContrast} extended this method to 3D point clouds. Crucially, their work focuses on generating representations of entire point clouds, i.e., they use a global approach. However, as Guerrero et al.~\cite{Guerrero-PCPNet} point out, normal estimation at a given point relies on the local structure of the point neighborhood rather than the global structure of the entire point cloud. This is also true for the problem of point cloud filtering~\cite{Rakotosaona-PCN} as effective filtering mechanisms must preserve sharp feature information locally. This motivates our approach of developing a patch-based contrastive learning mechanism where noise corruption of input patches is used as an augmentation to develop different views of the same underlying clean patch. Thereafter, we employ the Normalized Temperature-scaled Cross Entropy (NT-XEnt) loss function detailed in Sec.~\ref{sec:contrastive-learning} which promotes similarity of generated latent representations for a given positive pair of augmented patches. Inspired by the work of~\cite{Chen-SimCLR}, we do not explicitly sample negative pairs as the remaining augmented pairs within a batch can be used for this purpose. Furthermore, the goal of this contrastive process is to bring representations of patches of the same underlying clean structure closer together, which is unlike a triplet based learning process which simultaneously brings representations closer for similar patches while pushing away representations of dissimilar patches. 

\subsection{Joint normal estimation and filtering}
Normal estimation and point cloud filtering are two interconnected tasks. Accurately predicted normals are central to reliable point cloud filtering and surface reconstruction as mentioned by~\cite{Lu-Deep-Feature-Preserving,RIMLS-Oztireli}. In a similar manner, predicting normals on less noisy patches provide more accurate results, as opposed to noisier patches where outliers affect the final prediction~\cite{Guerrero-PCPNet,Lenssen-Deep-Iterative}. This motivates our joint normal estimation and filtering approach where our regression network estimates patch normals along with point displacements to filter central patch points. Thereafter, the estimated normals are used to further refine the final filtered position. This approach helps exploit the interlinked relationship between normal estimation and point cloud filtering and motivates our joint approach.

\begin{figure}[!ht]
\centering
\includegraphics[width=\linewidth]{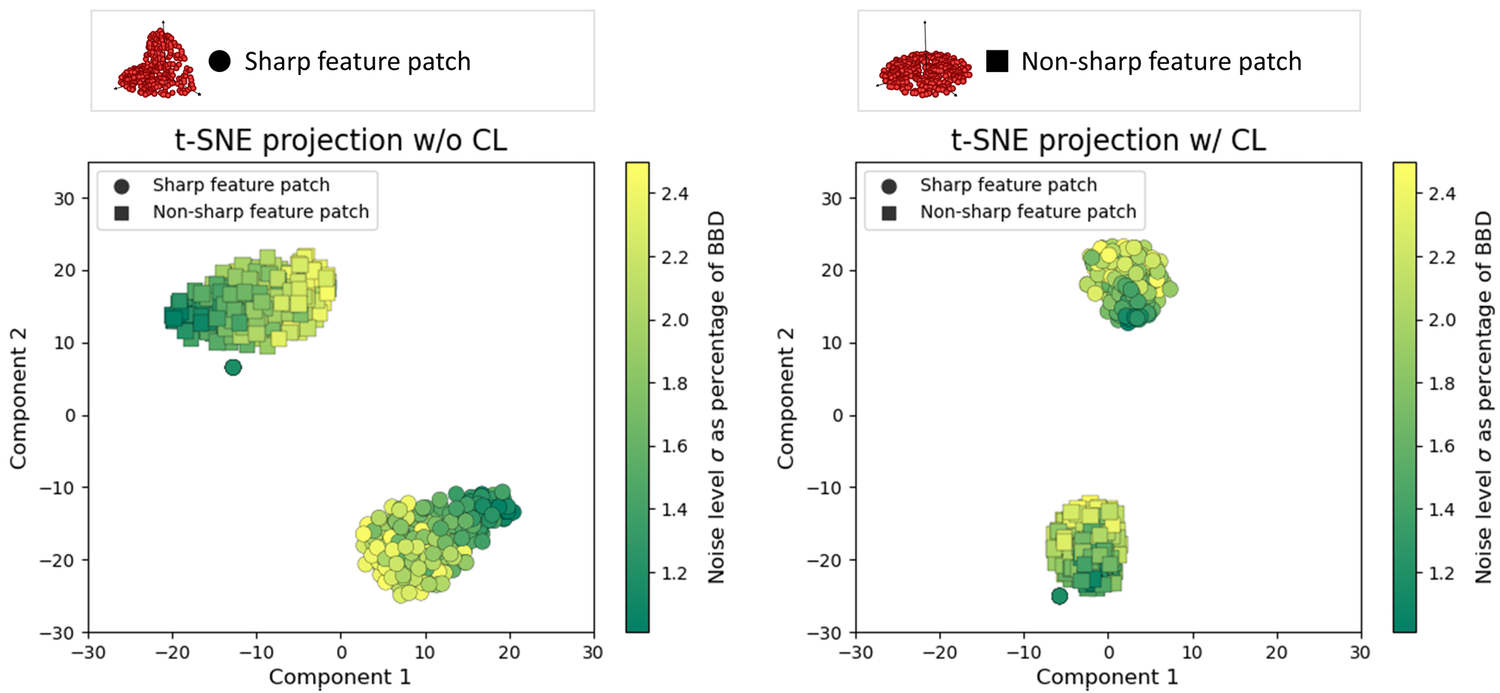}
\caption{
The impact of using a feature encoder pretrained using contrastive learning compared to a feature encoder trained from scratch. We consider t-SNE projections of latent representations of a sharp feature patch and a non-sharp feature patch at different Gaussian noise levels $\sigma$, w.r.t. the bounding box diagonal (BBD) of the clean point cloud. Respective clean patches are illustrated at the top.
}
\label{fig:two-patch-tsne}
\end{figure}

\subsection{Link between contrastive learning and regression tasks}
\begin{figure*}[!tp]
\centering
\includegraphics[width=6.8in]{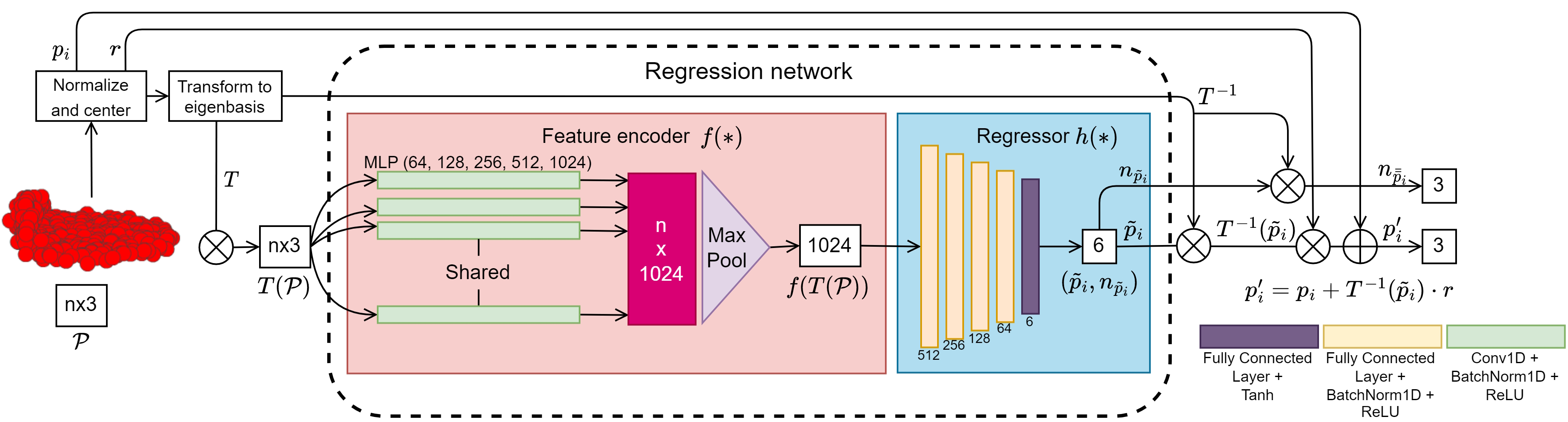}
\caption{Overview of the Regression network for outputting filtered points and normals. Feature encoder is taken from contrastive learning. }
\label{fig:full-network}
\end{figure*}

Feature encoders trained using contrastive learning are adept at generating similar representations of similar inputs and dissimilar representations of dissimilar inputs. Guided by the intuition that two noisy variants of the same underlying clean patch should generate similar latent representations, we develop a contrastive learning framework with noise corruption as an augmentation to train a feature encoder $f(\ast)$ (see Fig.~\ref{fig:contrastive-learning-framework}). This encoder is later used as part of a regression network (Fig.~\ref{fig:full-network}) to infer point normals and filtered displacements simultaneously. During training of the regression network, the pretrained feature encoder's weights are kept frozen and only the regressor $h(\ast)$ is trained. The pretrained feature encoder is robust to noise and generates representations that can be consumed more effectively by the regression network during training. Fig.~\ref{fig:two-patch-tsne} illustrates t-SNE projections of 250 noisy variants of a given sharp feature patch and 250 noisy variants of a non-sharp feature patch obtained from the Cube shape in our dataset. The corresponding clean patches are illustrated at the top of Fig.~\ref{fig:two-patch-tsne}. Each noisy variant contains Gaussian noise of standard deviation $\sigma$, ranging from $1.0\%$ to $2.5\%$ of the clean point cloud's bounding box diagonal. We observe that a feature encoder trained without contrastive learning generates latent representations that are less similar for differing noisy patch variants sharing a common underlying clean structure, for both the sharp feature and non-sharp feature patch. This is evident from Fig.~\ref{fig:two-patch-tsne} as low noise patch variants (dark green markers) have projections that are far apart from their respective higher noise counterparts (light green/yellow markers).

The feature encoder pretrained using contrastive learning, with only noise corruption as an augmentation, generates latent representations whose t-SNE projections are clustered more closely, indicating that representations are more similar even as noise increases. This is due to the contrastive pretraining that exploits noise corruption as an augmentation and ensures robustness to noise. As such, latent representations generated by the feature encoder are easier for the regression network to distinguish, even for high noise patches, as these representations are similar to that of the underlying clean counterparts. Thereby, the contrastive learning based pretraining facilitates our joint normal estimation and point cloud filtering method.

\section{Proposed Methodology}
\subsection{Overview}
We first introduce 3D patch-based contrastive learning to train a feature encoder capable of producing representations of point cloud patches (Fig.~\ref{fig:contrastive-learning-framework}). This feature encoder consists of a PointNet-like architecture, with 5 Conv1D layers and a global max pool layer that generates a 1024 dimensional representation of input point cloud patches. These representations are projected to a 256 dimensional vector by a projection head consisting of 3 fully connected layers. Once the feature encoder has been trained, we use it to generate latent representations of input patches for regression tasks. Next, we train a regression network (Fig.~\ref{fig:full-network}) to predict patch normals and displacements simultaneously (i.e., normal and displacement of the central point of a patch). The regressor consists of the pretrained feature encoder and a MLP of 5 fully connected layers that output the desired normals and displacements. During the testing phase, these displacements are added to the initial point cloud, which produces a filtered point cloud that is refined and becomes the input for the next iteration of inference.  

\subsection{Contrastive pair construction}
\label{sec:contrastive-pair-construction}
A clean point cloud consisting of $n$ points is described by $\mathbf{P}_s = \{p_i\ |\ p_i \in \mathbb{R}^3,\ i\ = 1, ..., n \}$ where $s=1,\ldots,M$ enumerates all training shapes. A noisy point cloud $\hat{\mathbf{P}}_s$ can, thereafter, be characterized by the addition of noise onto the clean point cloud
\begin{align} 
\label{eq:noisy-point-cloud}
\hat{\mathbf{P}}_s(\sigma) &= \mathbf{P}_s + \mathbf{N}(0, \sigma^2),
\end{align}
where $\mathbf{N}(0, \sigma^2)$ corresponds to additive Gaussian noise with a mean of $0$ and standard deviation of $\sigma$. For our training set, $\sigma$ takes on values of 0.25\%, 0.5\%, 1.0\%, 1.5\% and 2.5\% of the bounding box diagonal length of $\mathbf{P}_s$. 
For a given shape, the set of 6 variant point clouds (1 clean and 5 noisy), is given by
\begin{align}
    \mathbf{A}_s=\{\mathbf{P}_s, \hat{\mathbf{P}}_s(\sigma)\ |\ \sigma=0.25\%,\ldots,2.5\%\},
\end{align}
Patches sampled from point clouds in $\mathbf{A}_s$ are utilized in the contrastive learning process. 

An input patch centered at $p_i$, the $i^{th}$ point of a given point cloud $\mathbf{P}_s(\sigma_1) \in \mathbf{A}_s$, can be described by
\begin{align}
    \mathcal{P}=\{ p_j\ |\ p_j \in \mathbf{P}_s(\sigma_1) \land \norm{p_j - p_i}_2 < r_{\mathcal{P}}\},
\end{align}
We create a contrastive pair $(\mathcal{P}, \mathcal{Q})$ by randomly sampling another point cloud $\mathbf{P}_s(\sigma_2) \in \mathbf{A}_s$ such that,
\begin{align}
    \mathcal{Q} = \{ q_j\ |\ q_j \in \mathbf{P}_s(\sigma_2) \land \norm{q_j - p_i}_2 < r_{\mathcal{Q}}\},
\end{align}
Here, $\mathbf{P}_s(\sigma_1)$ and $\mathbf{P}_s(\sigma_2)$ correspond to two noisy variants of the same underlying clean point cloud and $\sigma_1$ and $\sigma_2$ are chosen randomly. The process of pairing $\mathcal{P}$ with $\mathcal{Q}$, where both share the same underlying clean patch structure, as both are subsets of $\mathbf{A}_s$ and are centered at $p_i$, is the first augmentation. We note that $\mathcal{P}$ may have a different patch radius to $\mathcal{Q}$ as the bounding box diagonal length of $\mathbf{P}_s(\sigma_1)$ differs from $\mathbf{P}_s(\sigma_2)$ unless $\sigma_1=\sigma_2$. The patch radii $r_{\mathcal{P}}$ and $r_{\mathcal{Q}}$ are taken to be 5\% of the respective bounding box diagonals. Patches are maintained at a fixed size to simplify the training procedure. Empirically, for each patch we sample 500 points. If the number of points is fewer, we increase it by copying random points in the patch and downsampling is applied otherwise. All patches are translated to the origin and normalized to [-1, 1], i.e., $\mathcal{P} = (\mathcal{P} - p_i)/r_{\mathcal{P}}$ and $\mathcal{Q} = (\mathcal{Q} - p_i)/r_{\mathcal{Q}}$.

For a batch of size $N$, we have a total of $2N$ augmented patches, i.e, we have input patches and their noise contrasted versions. The input patches are drawn from the training set $\mathbf{T}$, given by $\mathbf{T}=\bigcup_{s=1}^{M} \mathbf{A}_s$. All pairs $(\mathcal{P}_k, \mathcal{Q}_k)$ and $(\mathcal{Q}_k, \mathcal{P}_k)$, where $k=1,\ldots,N$, form positive contrastive pairs as these patches are drawn from point clouds of the same shape, the set $\mathbf{A}_s$, and correspond to the same central point $p_i$. Any pairs $(\mathcal{P}_k, \mathcal{Q}_l)$ or $(\mathcal{Q}_l, \mathcal{P}_k)$ where $k,l=1,\ldots,N \land k\neq l$ constitute negative pairs. For each augmented patch, there exists $1$ positive pair and $2(N-1)$ negative pairs. We do not pair the augmented patch with itself, i.e., we do not consider $(\mathcal{P}_k, \mathcal{P}_k)$ or $(\mathcal{Q}_k, \mathcal{Q}_k)$.

Positive contrastive pairs are put into the canonical basis by taking the inverse of the eigenvector matrix, obtained from the covariance matrix of $\mathcal{P}_k$, Eq.~\eqref{eq:patch-covariance-matrix}, and matrix multiplying both patches by it.
\begin{align} 
\label{eq:patch-covariance-matrix}
\mathcal{C}_{\mathcal{P}_k} = \frac{1}{|\mathcal{P}_k|}\sum_{p_j \in \mathcal{P}_k}(p_i-p_j)(p_i-p_j)^T,
\end{align}
Finally, we apply a second augmentation to $\mathcal{Q}_k$ by randomly rotating it around either the $x$, $y$, or $z$ axis by an angle $\theta \in \{0, \pi/12, \pi/6, \pi/4, \pi/3, \pi/2, 7\pi/12, 2\pi/3, 3\pi/4, 5\pi/6, \pi\}$. This second augmentation allows the network to learn patch similarity based on patch structure despite rotations and reinforces the effect of contrastive learning.

\subsection{Contrastive learning}
\label{sec:contrastive-learning}

Once contrastive pairs have been generated, we train our feature encoder $f(\ast)$ in a contrastive learning manner. As patches within a positive pair correspond to the same ground-truth patch, albeit with different levels of additive noise and arbitrary rotations applied to them, they should produce similar representations. As such, we adopt the Normalized Temperature-scaled Cross Entropy (NT-XEnt) loss~\cite{Chen-SimCLR} to achieve this. The loss for a positive pair $(\mathcal{P}_k,\mathcal{Q}_k)$ in the batch is given by
\begin{align} 
\label{eq:nt-xent-loss}
L_{\mathcal{P}_k,\mathcal{Q}_k} = -\log \frac{\exp{(\bm{z}_{\mathcal{P}_k} \cdot \bm{z}_{\mathcal{Q}_k})/\tau}}{\sum^{2N}_{l=1} \mathbbm{1}_{[k\neq l]}\exp{(\bm{z}_{\mathcal{P}_k} \cdot \bm{z}_{\mathcal{P}_l})/\tau}},
\end{align}
where $l=1,\ldots,2N$ enumerates over all projections. The function $\mathbbm{1}_{[k\neq l]} = 0$ if $k=l$ and $1$ otherwise. We empirically set $\tau=0.01$ based on results. In Eq.~\eqref{eq:nt-xent-loss}, the projection $\bm{z}_{\ast}=g(f(\ast))$ with $g(\ast)$ and $f(\ast)$ parameterized by the projection head and feature encoder, respectively. Finally, the contrastive loss for the entire batch is,
\begin{align} 
\label{eq:final-contrastive-loss}
L = \frac{1}{2N}\sum^{N}_{k=1} \left(L_{\mathcal{P}_k,\mathcal{Q}_k}+L_{\mathcal{Q}_k,\mathcal{P}_k}\right),
\end{align}
where $k=1,\ldots,N$ enumerates over all positive pairs. Fig.~\ref{fig:contrastive-learning-framework} shows how a single positive pair within a batch is processed by the feature encoder and projection head. All positive and negative pairs within the batch see the same copy of the feature encoder and projection head and all pairs are processed simultaneously. The larger the batch size, the greater the number of negative pairs available for the contrastive loss calculation. Once the training of the contrastive learning network has been completed, we discard the projection head and use the feature encoder as the backbone for our regression network. Representations generated by the feature encoder are consumed by the regressor, comprising 5 MLPs, and outputs a 2-tuple of displacement and normal vectors. During the training of the regressor, the weights for the feature encoder are kept frozen. The regression process is visualized in Fig.~\ref{fig:full-network}.

\subsection{Joint loss}
We propose a novel, joint position and normal based loss for training our regression network. This joint loss allows our network to, when trained, perform both a position update of a noisy point while also estimating its corresponding normal. Therefore, our regressor loss function comprises two main components: $L_{pos}$, the position loss and $L_{normal}$, the normal loss.

\textbf{Approximating the clean surface}. The position loss term is inspired by the work of~\cite{Rakotosaona-PCN}. We consider all points within the ground-truth patch $\mathcal{P}^{\star}$ and seek to minimize the squared $L_2$ norm instead of solely using a single fixed target.
\begin{equation}
L^1_{pos}(\tilde{p}_i, \mathcal{P}^{\star}) = \min_{p_j \in \mathcal{P}^{\star}}\norm{\tilde{p}_i - p_j}^2_2,
\end{equation}
where $\tilde{p}_i$ is the filtered point from the regressor given a noisy patch $\mathcal{P}$ centered at $p_i$, i.e., the $i^{th}$ point in $\mathcal{P}$. The ground truth patch $\mathcal{P}^{\star} = \{p_j\ |\ p_j \in \mathbf{P}_s \land \norm{p_j - p_i}_2 < r_{\mathcal{P}}\}$ is translated to the origin and normalized such that $\mathcal{P}^{\star} = (\mathcal{P}^{\star} - p_i)/r_{\mathcal{P}}$. This loss term allows the regressor to filter the noisy central point back to the clean patch surface. However, this does not ensure that the filtered point is centered within the ground truth patch which leads to unwanted clustering of filtered points.

\textbf{Ensuring regular distribution of points}. To avoid unwanted point clustering, we employ a regularization term which promotes the centering of filtered points within ground truth patches. By ensuring that filtered points lie close to the central points of their corresponding ground truth patches, we recover a regular distribution of points within the filtered point cloud.
\begin{equation}
\label{eq:regularization-term}
L^2_{pos}(\tilde{p}_i, \mathcal{P}^{\star}) = \max_{p_j \in \mathcal{P}^{\star}}\norm{\tilde{p}_i - p_j}^2_2,
\end{equation}
Intuitively, if the filtered point lies away from the ground truth patch center, this leads to a larger penalization due to Eq.~\eqref{eq:regularization-term}. For example, given a circular patch with radius $r$, the minimum value of Eq.~\eqref{eq:regularization-term} is $r^2$ where the inferred point lies at the center of the ground truth patch and the furthest point is $r$ away. $L^1_{pos}$ and $L^2_{pos}$ form the position based loss contribution to the final loss
\begin{equation}
\label{eq:position-loss}
L_{pos} = (1-\beta) L^1_{pos} + \beta L^2_{pos},
\end{equation}
where $\beta$ is the parameter that controls the regularization term's contribution to the position loss. We empirically set it to $0.01$, as we notice that a large contribution affects the convergence of the final loss function. 

\textbf{Normal estimation}. We then develop a relationship between a regressed position and its normal. Intuitively, if the ground-truth patch point, which minimizes the squared $L_2$ norm between positions, corresponds to the true position of the filtered point, then that point's normal should correspond to the true normal of the filtered point. If the ground-truth patch's central point which minimizes the squared $L_2$ distance is given by
\begin{equation}
p^{*}_j = \argmin_{p_j \in \mathcal{P}^{\star}}\left(\norm{\tilde{p}_i - p_j}^2_2\right),
\end{equation}
then the angle difference between the predicted normal and the ground-truth normal can be expressed in terms of cosine similarity between the two:
\begin{align}
\label{eq:normal-loss-cosine-similarity}
\cos(\theta) = n_{\tilde{p}_i}\cdot n_{p^{*}_j},
\end{align}
where $n_{\tilde{p}_i}$ and $n_{p^{*}_j}$ correspond to the normals at $\tilde{p}_i$ and $p^{*}_j$, respectively. This cosine term can now be used to construct our normal loss:
\begin{align}
\label{eq:normal-loss}
L_{normal} = 1-\left[\delta \cos(\theta)^2 + (1-\delta)\cos(\theta)^\gamma\right].
\end{align}

The loss function in Eq.~\eqref{eq:normal-loss} is a periodic function which penalizes $\theta$ values away from $0$ and $\pi$. Therefore, it encourages the predicted normal to be as close to the ground-truth normal as possible. It also assumes that the predicted normal with an angle difference of $\pi$ is equivalent to the ground-truth normal. The $\delta$ term, which is empirically set to $0.3$, serves to control the shape of loss function $L_{normal}$, wherein, angle differences close to $\pi/2$ are heavily penalized. This penalization decreases closer to $0$ and $\pi$. Finally, we express our joint loss as
\begin{equation}
\label{eq:joint-loss-function}
L_{final} = \alpha L_{pos} + (1-\alpha) L_{normal},
\end{equation}
where $\alpha$ controls the relative contributions of the position and normal losses to the final loss function. We empirically set $\alpha$ to $0.9$ as the emphasis for the regressor is denoising the point cloud iteratively. This in turn leads to better normal estimation results. 

\subsection{Alternative joint loss}
We also examine Eq.~\eqref{eq:alt-joint-loss-function}, a variant of the joint loss function, Eq.~\eqref{eq:joint-loss-function}, which utilizes the point-to-point correspondences between ground truth points $p^{\star}_i$ and filtered points $\tilde{p}_i$, i.e., using fixed ground truth targets.
\begin{align}
\label{eq:alt-position-loss}
L^{alt}_{pos} &= \norm{\tilde{p}_i - p^{\star}_i}^2_2, \\
\label{eq:alt-normal-loss}
L^{alt}_{normal} &= 1-[\delta\cos(\theta)^2 + (1-\delta)\cos(\theta)^\gamma], \\
\label{eq:alt-joint-loss-function}
L^{alt}_{final} &= \alpha L^{alt}_{pos} + (1-\alpha)L^{alt}_{normal},
\end{align}
where $\cos(\theta) = n_{\tilde{p}_i}\cdot n_{p^{\star}_i}$ with $n_{\tilde{p}_i}$ and $n_{p^{\star}_i}$ being the predicted normal and ground truth normal, respectively. As we regress the filtered point directly back to the ground truth central point, $L^{alt}_{pos}$ does not contain a repulsion term. The regressor trained with this loss function performs sub-optimally to that of the regressor trained using Eq.~\eqref{eq:joint-loss-function}. As noted in~\cite{Rakotosaona-PCN}, multiple clean points, within a given neighbourhood, may be perturbed in such a way as to result in the same noisy point. Therefore, the $L_2$ norm minimization between a filtered point and ground truth point, Eq.~\eqref{eq:alt-position-loss}, cannot successfully remove the noise component tangential to the surface and leads to a lower filtering performance. This motivates our use of Eq.~\eqref{eq:joint-loss-function} which regresses the filtered point back to the surface while ensuring it is centered as best possible within the ground truth patch. More details are given in Sec.~\ref{sec:alt-joint-loss}.

\subsection{Inference}

The regression network $h(\ast)$, trained based on the loss function defined by Eq.~\eqref{eq:joint-loss-function}, outputs a 6D vector or 2-tuple $(h_0, h_1)$ of 3D vectors. The first element corresponds to the displacement required to obtain the filtered point $p'_i$, and the second to the corresponding normal vector $\tilde{n}_i$. As these two vectors are in the space defined by the eigenvectors of the patch covariance matrix, they must first be transformed back to the original space. Subsequently, the filtered point is given by $p'_i = p_i + T^{-1}(h(f(T(\mathcal{P})))_0)\cdot r_{\mathcal{P}}$ with the original noisy point $p_i$. The normal vector in the original space is $n_{\Bar{\Bar{p}}_i} = T^{-1}(h(f(T(\mathcal{P}))))_1)$ where $f(\ast)$ is the feature encoder.

Given $p'_i$, we apply refinement during post-processing to obtain the final filtered point as suggested by~\cite{Rakotosaona-PCN,Lu-Low-Rank}. The first is a Taubin smoothing-like inflation step~\cite{Rakotosaona-PCN} and the second is the LRMA position update~\cite{Lu-Low-Rank} to combat shrinking of the point cloud and avoid incorrect displacement of points along the surface, respectively. The ablation study on the post-processing refinement is discussed in Section~\ref{sec:no-post-processing}. The new position after applying the inflation step is given by,
\begin{equation}
\label{eq:taubin-smoothing}
\bar{p}_i = p'_i - \frac{1}{|\mathcal{N}(p'_i)|}\sum_{p'_j\in \mathcal{N}(p'_i)}(p'_j-p_j),
\end{equation}
where we take $p'_j \in \mathcal{N}(p'_i)$ as the neighborhood of 100 filtered points in the vicinity of $p'_i$ and $p_j$ is the original point position before filtering. The LRMA position update yields our final filtered point. The position update is given by,
\begin{equation}
\label{eq:lr-position-update}
\Bar{\Bar{p}}_i = \bar{p}_i + \frac{1}{3|\mathcal{N}(\bar{p}_i)|}\sum_{\bar{p}_j\in \mathcal{N}(\bar{p}_i)}(\bar{p}_j-\bar{p}_i)(n^{T}_{\bar{p}_j}n_{\bar{p}_j} +n^{T}_{\bar{p}_i}n_{\bar{p}_i}),
\end{equation}
with $\mathcal{N}(\bar{p}_i)$ being the neighborhood of 20 points in the vicinity of $\bar{p}_i$.

\section{Experimental Results}
\subsection{Dataset}

\begin{figure}[!ht]
\centering
\includegraphics[width=0.49\linewidth]{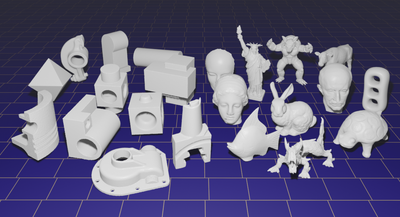}
\includegraphics[width=0.49\linewidth]{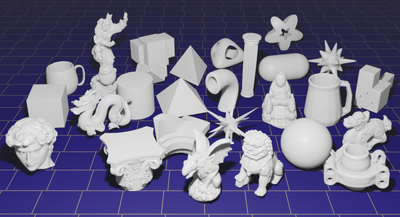}
\caption{Left: Meshes of synthetic point clouds used for training. Right: Meshes of synthetic point clouds used for validation (3 on the left) and for testing (23 on the right).}
\label{fig:mesh-train}
\end{figure}

Our training set consists of 22 synthetic point clouds (Fig.~\ref{fig:mesh-train}): 11 CAD shapes and 11 non-CAD shapes. The validation set consists of 3 shapes, 1 CAD and 2 non-CAD shapes, while the test set  consists of 23 shapes: 14 CAD and 9 non-CAD. Each shape is a point cloud of 100K points, which have been randomly sampled from their original surfaces. For training, we create 5 additional noisy variants of each training shape by adding Gaussian noise with standard deviations of 0.25\%, 0.5\%, 1.0\%, 1.5\% and 2.5\% of the clean point cloud's bounding box diagonal. These 6 variants (clean and 5 noise levels) for each shape give a total of 132 point clouds for training purposes. For validation, we consider 2 noisy variants of the initial 3 clean validation shapes, with 0.5\% and 1.0\% noise, resulting in a total of 6 validation point clouds. In the testing phase, we examine the robustness of our model at unseen noise levels by utilizing 0.6\%, 0.8\%, 1.1\%, 1.5\% and 2.0\% Gaussian noise for each test shape, yielding 115 test point clouds.

\begin{figure*}[!tp]
\centering
\includegraphics[width=5.4in]{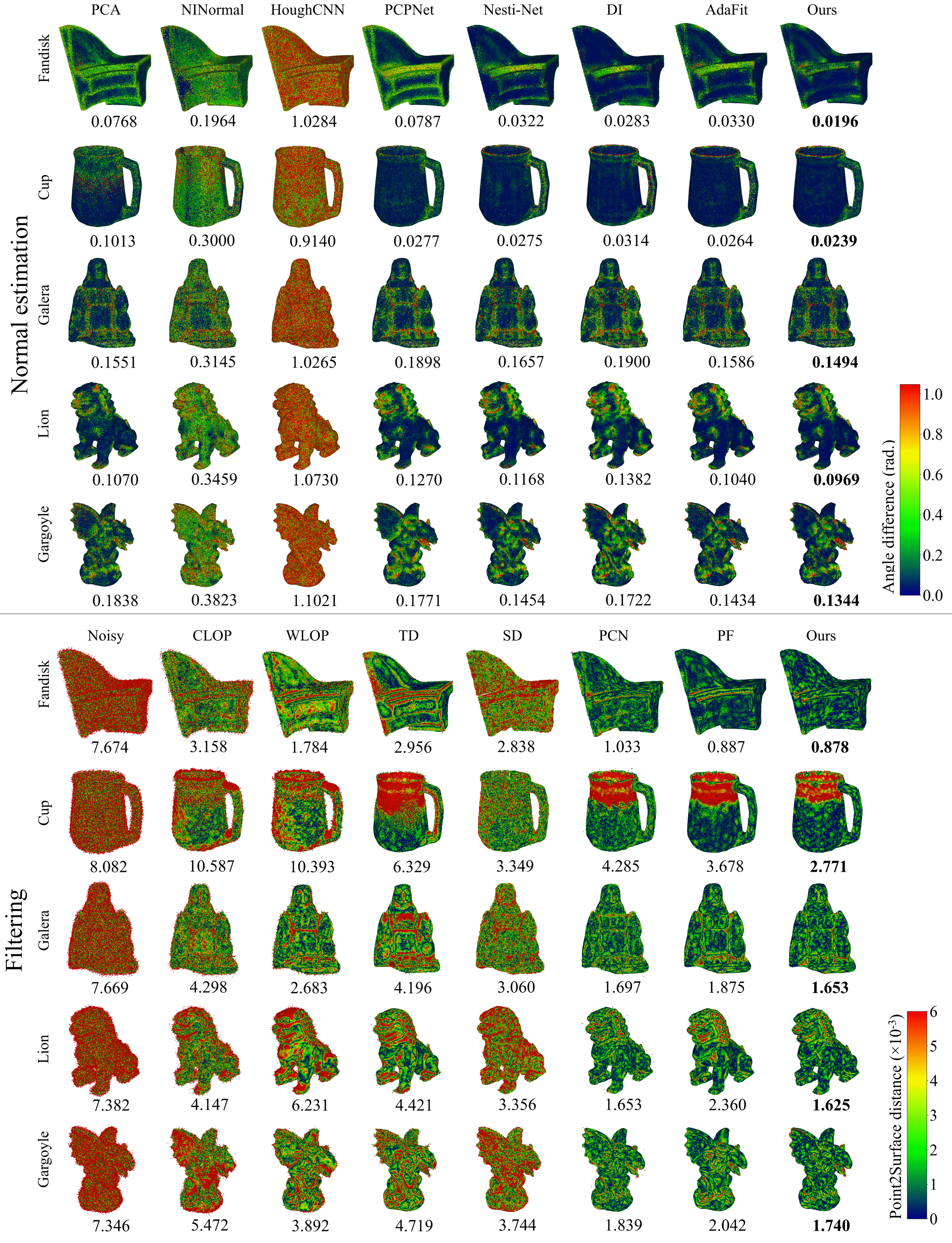}
\caption{Normal estimation (top half) and filtering (bottom half) results on shapes with 0.8\% Gaussian noise w.r.t. the bounding box diagonal. For normal estimation, the respective mean squared angular error (MSAE) is given below each shape and the heat map corresponds to the angle difference at each point. For  filtering, the Chamfer distance ($\times 10^{-5}$) is given below each shape and the heat map  corresponds to the scale normalized Point2Surface distance ($\times 10^{-3}$). }
\label{fig:all-synthetic-results}
\vspace{-0.8mm}
\end{figure*}

\textbf{Sharp features}. To evaluate performance at sharp features, we classify points within our synthetic dataset as feature and non-feature points. Please refer to the supplementary document for more details.

\subsection{Implementation}
The contrastive learning and regression networks are both trained on NVIDIA A100 GPUs using PyTorch 1.7.1 with CUDA 11.0. The contrastive learning network is trained for 150 epochs, with the Adam optimizer and a learning rate of $3\times 10^{-4}$. The regression network is trained for 30 epochs, utilizing the SGD optimizer with a learning rate of $1\times 10^{-2}$.

\subsection{Comparisons}
We compare our  method with state-of-the-art normal estimation and point cloud filtering methods. For normal estimation, we consider conventional PCA~\cite{Hoppe-PCA}, HoughCNN~\cite{Boulch-HoughCNN}, NINormal~\cite{Wang-NINormal}, PCPNet~\cite{Guerrero-PCPNet}, Nesti-Net~\cite{Ben-Shabat-Nesti-Net}, Deep Iterative (DI) \cite{Lenssen-Deep-Iterative} and AdaFit~\cite{Zhu-AdaFit} on our test set including synthetic and scanned point clouds. We do not compare with DeepFit~\cite{Ben-Shabat-DeepFit} as AdaFit is based on DeepFit and achieves better results. PCA requires the manual selection of neighborhood sizes for plane-fitting. For synthetic shapes, we utilize three different neighborhood sizes, i.e., 60 points for 0.6\% Gaussian noise, 150 points for 0.8\% Gaussian noise and 200 points for 1.1\%, 1.5\% and 2.0\% Gaussian noise. For scanned surfaces a neighborhood of 100 points is used. Our metric for comparison is the Mean Squared Angular Error (MSAE) where we calculate angle differences between ground truth normals $n_i$ and predicted normals $\tilde{n}_i$ and take the mean of their squares.

\midsepremove
\begin{table*}[!tp]
\centering
\caption{Normal estimation results on synthetic and scanned data given by the MSAE average. The MSAE for NINormal on scanned data is not given as it was not possible to test this method on input point clouds with over 100K points. The Gaussian noise standard deviation $\sigma$ is with respect to the bounding box diagonal of the clean point cloud. Top results in bold and second best results are underlined.}
\begin{tabular}{l|cccccccc}
\toprule
\multicolumn{1}{c|}{} & PCA   & NINormal & HoughCNN & PCPNet & Nesti-Net & DI    & AdaFit & Ours  \\
\midrule
Syn. ($\sigma$=0.6\%)       & 0.1637 & 0.2113 & 0.7418 & 0.1184 & \uline{0.1029} & 0.1037 & \textbf{0.0958} & 0.1038 \\
Syn. ($\sigma$=0.8\%)       & 0.1514 & 0.3277 & 1.0427 & 0.1414 & 0.1197 & 0.1246 & \uline{0.1143} & \textbf{0.1126} \\
Syn. ($\sigma$=1.1\%)       & 0.1954 & 0.4813 & 1.1595 & 0.1762 & \uline{0.1467} & 0.1539 & 0.1472 & \textbf{0.1268} \\
Syn. ($\sigma$=1.5\%)       & 0.2664 & 0.6160 & 1.1879 & 0.2174 & \uline{0.1827} & 0.1932 & 0.1914 & \textbf{0.1502} \\
Syn. ($\sigma$=2.0\%)       & 0.3632 & 0.7175 & 1.1853 & 0.2646 & \uline{0.2319} & 0.2405 & 0.2440 & \textbf{0.1951} \\
Syn. average              & 0.2280 & 0.4708 & 1.0634 & 0.1836 & \uline{0.1568} & 0.1632 & 0.1585 & \textbf{0.1377} \\
Sharp feat. ave.              & 0.4216 & 0.5872 & 1.0003 & 0.3947 & 0.3504 & 0.3913 & \uline{0.3445} & \textbf{0.3230} \\
Scanned average         & 0.1010 & --- & 0.4960 & 0.0550 & 0.0430 & \uline{0.0370} & \textbf{0.0250} & 0.0390
 \\
Overall average        & 0.2237 & 0.4550 & 1.0443 & 0.1793 & \uline{0.1530} & 0.1590 & 0.1540 & \textbf{0.1344}
 \\
\bottomrule
\end{tabular}
\label{tab:normal-estimation-all}
\end{table*}
\midsepdefault

\midsepremove
\begin{table*}[!tp]
\centering
\caption{Average Chamfer and Point2Surface distance results on synthetic and scanned data. The Gaussian noise standard deviation $\sigma$ is with respect to the bounding box diagonal of the clean point cloud. Top results in bold and second best results are underlined.}
\begin{tabular}{l|l|cccccccc}
\toprule
& & Noisy & CLOP  & WLOP  & TD & SD    & PCN   & PF             & Ours           \\
\midrule
\multirow{8}{*}{\pbox{100pt}{Chamfer dist. \\($\times10^{-5}$)}} & Syn. ($\sigma$=0.6\%) & 4.762 & 2.366 & 3.374 & 4.638 & 1.867 & 1.467 & \uline{1.444} & \textbf{1.241} \\
& Syn. ($\sigma$=0.8\%) & 7.517 & 4.163 & 3.921 & 4.038 & 3.127 & 1.752 & \uline{1.749} & \textbf{1.545} \\
& Syn. ($\sigma$=1.1\%) & 12.746 & 8.311 & 6.533 & 3.508 & 6.903 & 2.572 & \uline{2.355} & \textbf{1.987} \\
& Syn. ($\sigma$=1.5\%) & 21.637 & 16.13 & 14.24 & 7.472 & 14.301 & 5.787 & \uline{3.544} & \textbf{3.244} \\
& Syn. ($\sigma$=2.0\%) & 35.906 & 29.177 & 27.318 & 22.838 & 27.943 & 17.658 & \textbf{5.307} & \uline{8.055} \\
& Syn. average       & 16.514 & 12.029 & 11.077 & 8.499 & 10.828 & 5.847 & \textbf{2.880} & \uline{3.214} \\
& Scanned average         & 2.764 & 3.317 & 5.618 & 3.835 & 3.443 & 0.542 & \uline{0.482} & \textbf{0.472} \\
& Overall average           & 16.052 & 11.736 & 10.894 & 8.342 & 10.58 & 5.669 & \textbf{2.799} & \uline{3.122} \\
\midrule
\multirow{9}{*}{\pbox{100pt}{Point2Surf. \\ dist. ($\times10^{-3}$)}} & Syn. ($\sigma$=0.6\%) & 4.658 & 2.392 & 2.423 & 3.195 & 2.262 & 1.380 & \uline{1.141} & \textbf{1.067} \\
& Syn. ($\sigma$=0.8\%) & 6.143 & 3.730 & 3.052 & 3.078 & 3.520 & 1.592 & \uline{1.386} & \textbf{1.324} \\
& Syn. ($\sigma$=1.1\%) & 8.306 & 5.976 & 4.810 & 3.063 & 6.009 & 2.108 & \uline{1.810} & \textbf{1.807} \\
& Syn. ($\sigma$=1.5\%) & 11.073 & 8.959 & 8.418 & 5.471 & 9.259 & 3.100 & \textbf{2.472} & \uline{2.937} \\
& Syn. ($\sigma$=2.0\%) & 14.432 & 12.685 & 12.460 & 11.282 & 13.315 & 5.124 & \textbf{3.440} & \uline{5.020} \\
& Syn. average       & 8.922 & 6.748 & 6.233 & 5.218 & 6.873 & 2.661 & \textbf{2.050} & \uline{2.431} \\
& Sharp feat. ave.       & 9.082 & 6.867 & 6.324 & 6.409 & 6.893 & 3.366 & \textbf{2.339} & \uline{2.887} \\
& Scanned average       & 3.795 & 3.410 & 4.037 & 2.990 & 3.957 & 1.000 & \uline{0.722} & \textbf{0.705} \\
& Overall average      & 8.750 & 6.636 & 6.159 & 5.143 & 6.775 & 2.605 & \textbf{2.005} & \uline{2.373} \\
\bottomrule
\end{tabular}
\label{tab:filtering-all}
\end{table*}
\midsepdefault

For point cloud filtering, we compare with conventional methods CLOP~\cite{Preiner-CLOP} and WLOP~\cite{Huang-WLOP} and deep learning methods TotalDenoising (TD)~\cite{Hermosilla-Total-Denoising}, ScoreDenoise (SD)~\cite{Luo-Score-Based-Denoising}, PointCleanNet~\cite{Rakotosaona-PCN} and Pointfilter~\cite{Zhang-Pointfilter} on two metrics, the Chamfer distance (CD) and the Point2Surface distance (P2S)~\cite{Li-Dis-PU}. The Chamfer distance between a ground truth point cloud $\mathbf{P}_s$ and a filtered point cloud $\mathbf{\tilde{P}}$ is defined by Eq.~\eqref{eq:chamfer}. The first term provides a measure of the distance from filtered points to the ground truth surface while the second provides a measure of the relative even distribution of filtered points w.r.t. the clean point cloud. P2S measures the average distance between filtered points and the reconstructed ground truth mesh. All learning based methods are retrained on our dataset.
\begin{align}
    C(\mathbf{P}_s, \mathbf{\tilde{P}}) = &\frac{1}{\vert\mathbf{\tilde{P}}\vert}\sum_{p_i\in \mathbf{\tilde{P}}} \min_{p_j\in \mathbf{P}_s}\norm{p_i-p_j}^2_2 \notag\\ &+ \frac{1}{\vert\mathbf{P}_s\vert}\sum_{p_j\in \mathbf{P}_s} \min_{p_i\in \mathbf{\tilde{P}}}\norm{p_j-p_i}^2_2
    \label{eq:chamfer}
\end{align}

\textbf{Evaluation at sharp features}. 
We take the MSAE for sharp feature points as a measure of normal estimation accuracy and the P2S distance for these points as a measure of filtering accuracy. Tables~\ref{tab:normal-estimation-all},~\ref{tab:filtering-all},~\ref{tab:normal-estimation-additional} and~\ref{tab:filtering-additional} give the MSAE and P2S averages for each method on our dataset.

\subsection{Performance on synthetic data} 
Tables~\ref{tab:normal-estimation-all} and~\ref{tab:filtering-normal-individual} and the top half of Fig.~\ref{fig:all-synthetic-results} demonstrate normal estimation results on shapes with Gaussian noise. For the normal estimation task, our method outperforms all other methods including ones which employ larger networks, such as Nesti-Net, and recovers accurate normals in the presence of noise. AdaFit offers competitive results at lower noise levels but performs sub-optimally at higher noise. Another key attribute is its ability to accurately predict normals at sharp features. Both these attributes of the network can be seen by its performance on the Fandisk shape. Overall, it has the lowest MSAE at sharp feature points and shows higher robustness to noise. Contrastive learning facilitates this as patches from anisotropic surfaces are trained to produce distinct feature representations which can be better distinguished by the regressor. Additionally, as positive contrastive pairs share the same underlying clean surface, representations of a given patch at differing noise levels remain similar. This allows the regressor to reliably estimate normals as noise increases.

\midsepremove
\begin{table*}[!ht]
\centering
\caption{Individual MSAE and Chamfer distance values for shapes presented in Fig.~\ref{fig:all-synthetic-results}, at Gaussian noise levels of standard deviation $\sigma$, with respect to the bounding box diagonal of the clean point cloud.}
\begin{tabular}{c|c|ccccc|cccc}
\toprule
\multirow{2}{*}{Shape} & \multicolumn{1}{c|}{\multirow{2}{*}{$\sigma$}} & \multicolumn{5}{c|}{Normal estimation - MSAE} & \multicolumn{4}{c}{Filtering - Chamfer distance ($\times10^{-5}$)} \\
\cmidrule{3-11}
 & \multicolumn{1}{c|}{} & PCPNet & Nesti-Net & DI & AdaFit & Ours & Noisy & PCN & PF & Ours \\
 \midrule
\multirow{5}{*}{Fandisk} & 0.6\% & 0.0615 & 0.0205 & \textbf{0.0148} & 0.0197 & \uline{0.0159} & 4.735 & 0.896 & \uline{0.748} & \textbf{0.747} \\
 & 0.8\% & 0.0787 & 0.0322 & \uline{0.0283} & 0.0329 & \textbf{0.0196} & 7.674 & 1.033 & \uline{0.887} & \textbf{0.878} \\
 & 1.1\% & 0.1005 & \uline{0.0534} & 0.0542 & 0.0563 & \textbf{0.0266} & 13.232 & 1.546 & \uline{1.246} & \textbf{1.244} \\
 & 1.5\% & 0.1250 & \uline{0.0708} & 0.0828 & 0.0948 & \textbf{0.0442} & 22.610 & 3.520 & \textbf{1.678} & \uline{2.285} \\
 & 2.0\% & 0.1498 & \uline{0.0945} & 0.1152 & 0.1366 & \textbf{0.0716} & 37.450 & 13.952 & \textbf{2.916} & \uline{6.613} \\
\midrule
\multirow{5}{*}{Cup} & 0.6\% & 0.0194 & 0.0218 & 0.0262 & \textbf{0.0198} & \uline{0.0213} & 5.276 & 3.488 & \uline{2.384} & \textbf{2.117} \\
 & 0.8\% & 0.0277 & 0.0275 & 0.0314 & \uline{0.0264} & \textbf{0.0239} & 8.082 & 4.285 & \uline{3.678} & \textbf{2.771} \\
 & 1.1\% & 0.0451 & 0.0399 & \uline{0.0389} & 0.0446 & \textbf{0.0287} & 12.961 & 5.609 & \uline{5.228} & \textbf{2.957} \\
 & 1.5\% & 0.0773 & 0.0645 & \uline{0.0518} & 0.0749 & \textbf{0.0390} & 21.329 & 10.267 & \uline{8.462} & \textbf{2.995} \\
 & 2.0\% & 0.1153 & 0.1108 & \uline{0.0755} & 0.1027 & \textbf{0.0640} & 34.351 & 22.665 & \uline{10.917} & \textbf{5.057} \\
\midrule
\multirow{5}{*}{Galera} & 0.6\% & 0.1504 & 0.1357 & 0.1469 & \textbf{0.1256} & \uline{0.1324} & 4.924 & \uline{1.422} & 1.516 & \textbf{1.334} \\
 & 0.8\% & 0.1898 & 0.1657 & 0.1899 & \uline{0.1586} & \textbf{0.1494} & 7.669 & \uline{1.697} & 1.875 & \textbf{1.653} \\
 & 1.1\% & 0.2416 & \uline{0.2085} & 0.2472 & 0.2091 & \textbf{0.1747} & 12.922 & \uline{2.430} & 2.478 & \textbf{2.128} \\
 & 1.5\% & 0.2904 & \uline{0.2609} & 0.2947 & 0.2670 & \textbf{0.2166} & 21.713 & 5.930 & \uline{3.493} & \textbf{3.324} \\
 & 2.0\% & 0.3403 & \uline{0.3136} & 0.3441 & 0.3219 & \textbf{0.2775} & 35.215 & 19.101 & \textbf{6.112} & \uline{6.879} \\
\midrule
\multirow{5}{*}{Lion} & 0.6\% & 0.0957 & 0.0896 & 0.0989 & \textbf{0.0741} & \uline{0.0798} & 4.691 & \uline{1.309} & 1.768 & \textbf{1.219} \\
 & 0.8\% & 0.1270 & 0.1168 & 0.1382 & \uline{0.1040} & \textbf{0.0969} & 7.382 & \uline{1.653} & 2.360 & \textbf{1.625} \\
 & 1.1\% & 0.1707 & 0.1570 & 0.1786 & \uline{0.1481} & \textbf{0.1262} & 12.489 & \uline{2.567} & 3.160 & \textbf{2.273} \\
 & 1.5\% & 0.2205 & 0.2053 & 0.2185 & \uline{0.2022} & \textbf{0.1684} & 21.135 & 6.105 & \uline{4.130} & \textbf{3.606} \\
 & 2.0\% & 0.2755 & 0.2630 & 0.2718 & \uline{0.2595} & \textbf{0.2279} & 34.894 & 19.627 & \textbf{6.181} & \uline{8.796} \\
\midrule
\multirow{5}{*}{Gargoyle} & 0.6\% & 0.1305 & \uline{0.1113} & 0.1250 & \textbf{0.1038} & 0.1134 & 4.710 & \uline{1.518} & 1.661 & \textbf{1.345} \\
 & 0.8\% & 0.1771 & 0.1454 & 0.1722 & \uline{0.1434} & \textbf{0.1344} & 7.346 & \uline{1.839} & 2.042 & \textbf{1.74} \\
 & 1.1\% & 0.2445 & \uline{0.2037} & 0.2340 & 0.2159 & \textbf{0.1753} & 12.16 & \uline{2.968} & 3.034 & \textbf{2.441} \\
 & 1.5\% & 0.3077 & \uline{0.2672} & 0.2953 & 0.2865 & \textbf{0.2253} & 19.903 & 6.594 & \uline{5.179} & \textbf{3.902} \\
 & 2.0\% & 0.3725 & \uline{0.3383} & 0.3616 & 0.3523 & \textbf{0.3019} & 32.178 & 19.408 & \textbf{8.000} & \uline{8.597} \\
\bottomrule
\end{tabular}
\label{tab:filtering-normal-individual}
\end{table*}
\midsepdefault

\begin{figure}[!tp]
\centering
\includegraphics[width=3.5in]{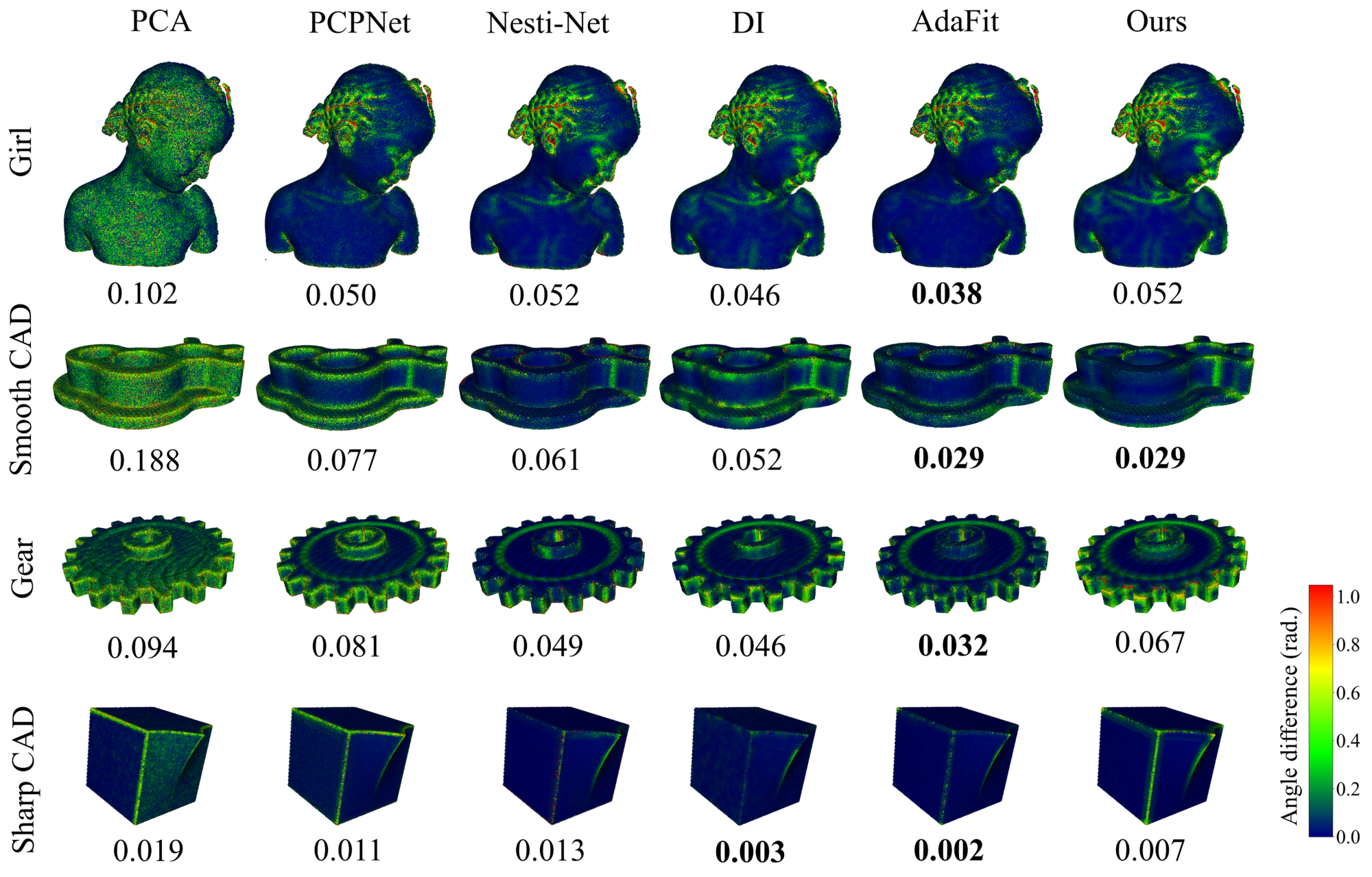}
\caption{Normal estimation results on scanned shapes.}
\label{fig:normal-estimation-all-scanned}
\end{figure}

\begin{figure}[!tp]
\centering
\includegraphics[width=3.5in]{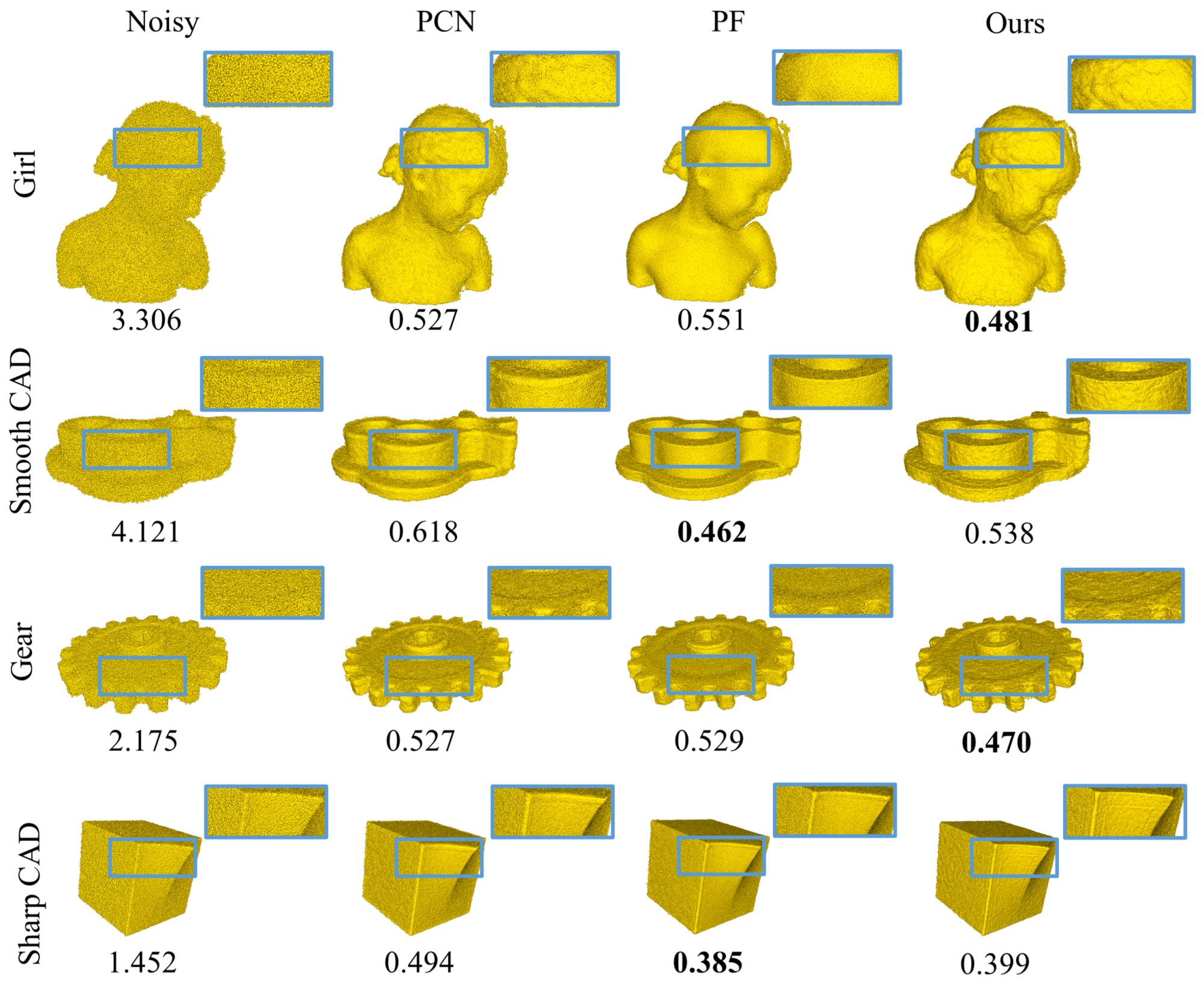}
\caption{Visual filtering results on scanned shapes.}
\label{fig:filtering-all-scan}
\end{figure}

\begin{figure*}[!tp]
\centering
\includegraphics[width=6.8in]{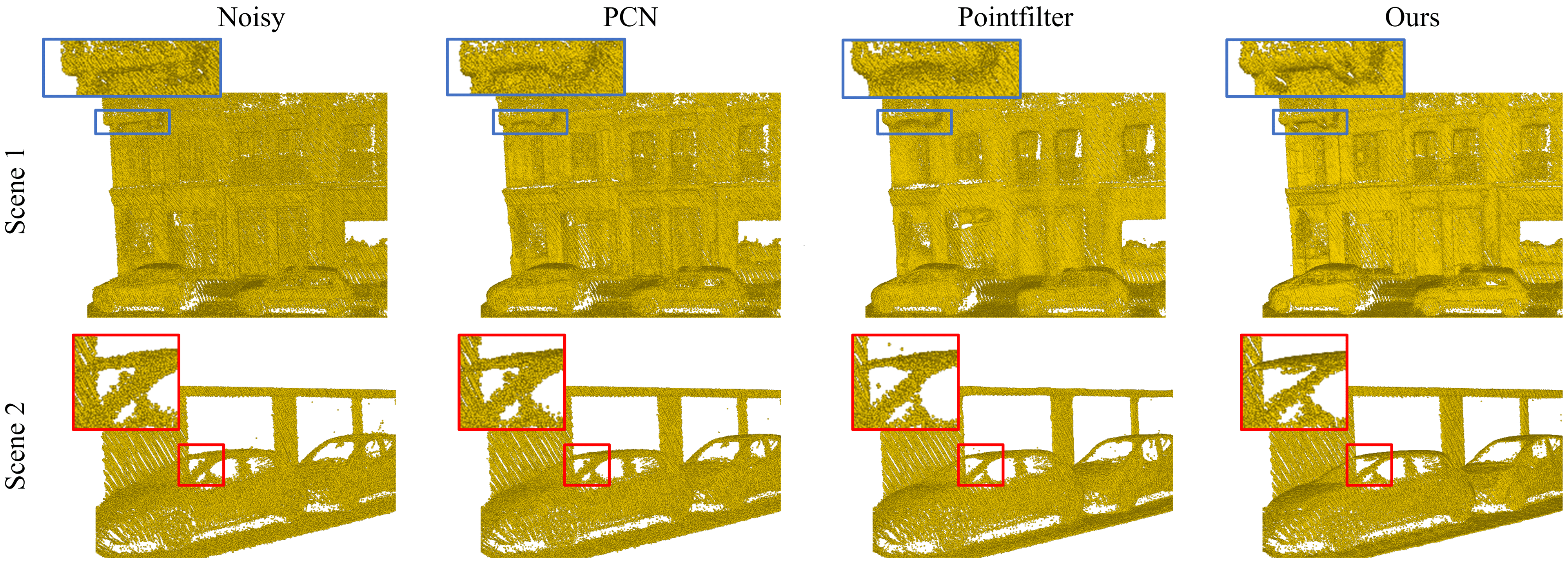}
\caption{Visual filtering results for the two scenes extracted from the Rue-Madame data. Our method recovers sharp features and fine details of vehicles and building facades while other methods perform sub-optimally.}
\label{fig:filtering-rm-scan}
\end{figure*}

Table~\ref{tab:filtering-all} and~\ref{tab:filtering-normal-individual} and the bottom half of Fig.~\ref{fig:all-synthetic-results} demonstrate filtering results on shapes with Gaussian noise. A core attribute of our method is that it generalises well between both CAD-like and non-CAD like shapes. It is able to recover the sharp feature information of CAD shapes such as Fandisk while also achieving the highest accuracy on non-CAD shapes such as Galera and Gargoyle. Furthermore, on complex shapes such as Cup, our method outperforms others at filtering noisy points along closely neighboring surfaces. Pointfilter specializes in the point cloud filtering task and it only outperforms our method on both metrics at the highest noise level. Although, at higher noise, Pointfilter has slightly lower Point2Surface distances, the higher Chamfer distance indicates a relatively uneven distribution of points along the filtered surface. This is due to the fact that there are multiple displacements along which a point may return to the underlying clean surface. Therefore, even if the point is returned to the underlying surface, it may be incorrectly positioned along the surface. Our method reduces this longitudinal jitter by using estimated normals in conjunction with the LRMA position update scheme of~\cite{Lu-Low-Rank}. This takes regressed point positions, the output from our network, as input and updates them based on the neighboring points' positions and normal information from the previous iteration. Thereby, we are able to optimize both the position and normal estimation predictions to generate a filtered point cloud which more accurately represents the original clean point cloud. Additional visual results are given in the supplementary document.

\subsection{Performance on scanned data} 
Next, we look at the performance of each method on the scanned dataset which comprises 3 CAD-like scans and 1 non-CAD like scan where the ground-truth normal and position information are known. These scans are obtained using the virtual scanner introduced by Yu et al~\cite{Yu-EC-Net}. Additionally, we present comparisons on the Kinect v1 and Kinect v2 datasets, introduced by Wang, Liu and Tong~\cite{Wang-Kinect}, comprising 71 and 72 scans respectively, obtained using Kinect v1 and v2 sensors. As these scans have no ground truth normal information, we only present filtering results. We also test the learning based filtering methods on two scans extracted from the Paris-Rue-Madame database~\cite{Serna-Rue-Madame}. They are scans taken from Rue Madame, a street in the 6\textsuperscript{th} district of Paris. These scans capture many intricate details like the street's building facades and parked vehicles and also provide a good example of real-world noise that is encountered when obtaining such scans. Finally, we provide results on two scans extracted from the Kitti-360 dataset~\cite{Liao-Kitti360} in the supplementary document. They capture scenes in several suburbs of Karlsruhe, Germany, using  Velodyne HDL-64E sensors. As the Rue Madame and Kitti-360 scans have no ground truth, only visual results are presented. Tables~\ref{tab:normal-estimation-all} and~\ref{tab:filtering-all} illustrate quantitative normal estimation and filtering results on the scanned data. AdaFit outperforms other methods in estimating normals on scanned data, but is less robust to high noise. Our method, on average, performs better than other methods and remains competitive at estimating normals near sharp features, as shown in Fig.~\ref{fig:normal-estimation-all-scanned}. 

\midsepremove
\begin{table}[!ht]
\centering
\caption{Filtering results on the Kinect v1 and Kinect v2 datasets. Average Chamfer distance ($\times10^{-5}$) and Point2Surface distance ($\times10^{-3}$) values are presented for each method.}
\begin{tabular}{l|l|cccccccc}
\toprule
& & Noisy & PCN & PF & Ours \\
\midrule
\multirow{2}{*}{\pbox{100pt}{Kinect v1}} & CD & 14.489 & 13.469 & \uline{12.623} & \textbf{12.083} \\
& P2S & 6.272 & 5.763 & \uline{5.029} & \textbf{4.947} \\
\midrule
\multirow{2}{*}{\pbox{100pt}{Kinect v2}} & CD & 22.633 & 21.985 &  \uline{20.174} & \textbf{18.785} \\
& P2S & 7.505 & 7.269 & \uline{6.265} & \textbf{5.889} \\
\bottomrule
\end{tabular}
\label{tab:kinect-data}
\end{table}
\midsepdefault

Among filtering methods, our method shows an advantage in recovering sharp feature information as well as fine details, such as the braids in the girl's hair in Fig.~\ref{fig:filtering-all-scan}. Our method also performs optimally on the Kinect v1 and Kinect v2 datasets, as demonstrated by Table~\ref{tab:kinect-data}. We outperform other methods on both the Chamfer distance and Point2Surface metrics, indicating its ability to approximate the surface and generate a regular distribution of points. Visual results on the Kinect data are presented in the supplementary document. Furthermore, Fig.~\ref{fig:filtering-rm-scan} demonstrates our method's ability to filter complex scenes. Results on Scene 1 and Scene 2 demonstrate our method's ability to maintain sharp features, in the presence of real-world noisy artifacts, while other methods such as PointCleanNet (PCN) and Pointfilter (PF)~\cite{Rakotosaona-PCN,Zhang-Pointfilter} tend to smear feature information. For example, in Scene 1, the outlines of doors and windows are filtered accurately by our method while smoothing planar surfaces (surfaces of doors, walls). Fine details such as the headlights of vehicles and the outlines of tire rims are also recovered. Conversely, Pointfilter smears such details while PointCleanNet only partially filters them and is not successful in smoothing planar surfaces. In Scene 2, we demonstrate an ability to reliably filter details of vehicles including side-mirrors and windows while other methods perform sub-optimally.

Based on these results, we see that our method is very competitive in filtering while holding a clear advantage in normal estimation. On average (Tables~\ref{tab:normal-estimation-all} and~\ref{tab:filtering-all}), our method outperforms other methods in both tasks.

\subsection{Performance on shapes with varying density and different noise patterns} 
The synthetic data results presented in Tables~\ref{tab:normal-estimation-all} and~\ref{tab:filtering-all} correspond to point clouds generated by uniformly sampling their original meshes, with Gaussian random noise subsequently added to them. We also evaluate normal estimation and filtering methods on uniformly sampled point clouds with an impulsive noise pattern and on point clouds sampled with varying density. To generate point clouds with an impulsive noise pattern, we apply Gaussian noise of $\sigma=1.5\%$, with respect to the bounding box diagonal, to 30\% of points in each clean point cloud within the test set. To obtain varying density point clouds, two different sampling regimes, gradient and striped, are applied to the original test meshes. Thereafter, Gaussian noise of $\sigma=0.8\%$ is applied to these point clouds. Further details on varying density point clouds can be found in the supplementary document. When evaluating the performance of methods on test  point clouds with varying density and different noise patterns, we use models trained on our synthetic dataset comprising of point clouds with Gaussian noise. Table~\ref{tab:normal-estimation-additional} and Table~\ref{tab:filtering-additional} demonstrate our method's ability to generalize well on point clouds with varying density for normal estimation while performing competitively in filtering. For point clouds with an impulse noise pattern, we perform optimally at the filtering task.

\midsepremove
\begin{table}[!tp]
\setlength{\tabcolsep}{2pt}
\centering
\caption{MSAE results on point clouds with varying density (VD) and Impulsive Noise (IN). The Gaussian noise standard deviation $\sigma$ is with respect to the bounding box diagonal of the clean point cloud.}
\begin{tabular}{l|cccccccc}
\toprule
\multicolumn{1}{c|}{} & PCA & PCPNet & Nesti-Net & DI & AdaFit & Ours \\
\midrule
VD ($\sigma$=0.8\%) & 0.143 & 0.102 & 0.098 & 0.089 & \uline{0.083} & \textbf{0.075} \\
VD sharp feat. ave. & 0.298 & 0.294 & 0.273 & 0.278 & \uline{0.247} & \textbf{0.242} \\
IN ($\sigma$=1.5\%) & 0.118 & 0.125 & 0.106 & \textbf{0.077} & \uline{0.104} & 0.111 \\
IN sharp feat. ave. & 0.334 & 0.329 & \uline{0.275} & \textbf{0.208} & 0.304 & 0.309 \\
\bottomrule
\end{tabular}
\label{tab:normal-estimation-additional}
\end{table}
\midsepdefault

\midsepremove
\begin{table}[!tp]
\centering
\caption{Average Chamfer and Point2Surface distance results on point clouds with varying density (VD) and Impulsive Noise (IN). The Gaussian noise standard deviation $\sigma$ is with respect to the bounding box diagonal of the clean point cloud.}
\begin{tabular}{l|l|cccccccc}
\toprule
& & Noisy & PCN & PF & Ours \\
\midrule
\multirow{2}{*}{\pbox{100pt}{CD \\($\times10^{-5}$)}} & VD ($\sigma$=0.8\%) & 7.560 & 1.730 & \uline{1.680} & \textbf{1.530} \\
& IN ($\sigma$=1.5\%) & 6.366 & 3.279 & \uline{1.001} & \textbf{0.563} \\
\midrule
\multirow{4}{*}{\pbox{100pt}{P2S \\($\times10^{-3}$)}} & VD ($\sigma$=0.8\%) & 6.272 & 1.844 & \textbf{1.564} &  \uline{1.637} \\
& VD sharp feat. ave. & 6.380 & 2.598 & \textbf{2.336} & \uline{2.338} \\
& IN ($\sigma$=1.5\%) & 3.411 & 1.862 & \uline{1.451} & \textbf{0.897} \\
& IN sharp feat. ave. & 3.568 & 2.993 & \uline{2.597} & \textbf{1.383} \\
\bottomrule
\end{tabular}
\label{tab:filtering-additional}
\end{table}
\midsepdefault

\subsection{Comparative runtimes}
Finally, we look at the comparative runtimes for the different methods among the best performing normal estimation and filtering methods, respectively. The runtimes for normal estimation are 20.9, 2.0 minutes/100K points for Nesti-Net and AdaFit while for filtering, PointCleanNet and Pointfilter take 27.2, 1.8 minutes/100K points, respectively. Our method, by comparison, takes 4.9 minutes/100K points for the combined normal estimation and filtering tasks.

\section{Ablation Study}
\label{sec:ablation-study}
We empirically found optimal values for the three main hyper-parameters, which are $\alpha=0.9$, $\gamma=12$ and the contrastive batch size of 512. Please refer to Sec.~\ref{sec:ablation-hyperparams} for details.

\midsepremove
\begin{table}[!ht]
\centering
\caption{MSAE and Chamfer distance results on the validation set when only one task, normal estimation ($\gamma$=12, CBS=512) or filtering (CBS=512), is performed as opposed to when they are jointly performed ($\alpha=0.9$, $\gamma$=12, CBS=512).}
\begin{tabular}{l|ccc}
\toprule
& Normal estimation & Filtering & Both \\
\midrule
MSAE & \uline{0.052} & --- & \textbf{0.034} \\
\midrule
CD $(\times10^{-5})$ & --- & \uline{1.609} & \textbf{1.431} \\
\bottomrule
\end{tabular}
\label{tab:ablation-one-task}
\end{table}
\midsepdefault

\subsection{Merely normal estimation or filtering} 
We test the variants with merely normal estimation or point cloud filtering. Table~\ref{tab:ablation-one-task} shows the results of merely normal estimation, merely point cloud filtering, and the proposed joint approach on the validation set. We see that our joint approach achieves best quantitative results in terms of both MSAE and Chamfer distance, confirming the effectiveness of our joint method. 

\subsection{Alternative joint loss} 
\label{sec:alt-joint-loss}
Next, we look at the performance of the alternative loss function, Eq.~\eqref{eq:alt-joint-loss-function}, which considers the point-to-point correspondences between ground truth points $p^{\star}_i$ and filtered points $\tilde{p}_i$, i.e., using fixed ground truth targets. We see poorer results for this alternative loss function on the validation set, i.e., MSAE: $0.038$ vs $0.034$, Chamfer distance: $1.58\times10^{-5}$ vs $1.43\times10^{-5}$. The latter results correspond to the original joint loss in Eq.~\eqref{eq:joint-loss-function}. As expected, the original joint loss produces filtered point clouds closer to the corresponding ground truth point clouds, with smaller Chamfer distances between them while the $L_2$ norm minimization of the alternative joint loss function, Eq.~\eqref{eq:alt-joint-loss-function}, is less successful at the filtering task.

\subsection{Effect of 3D patch based contrastive learning} 
\begin{figure}[!ht]
\centering
\includegraphics[width=3.5in]{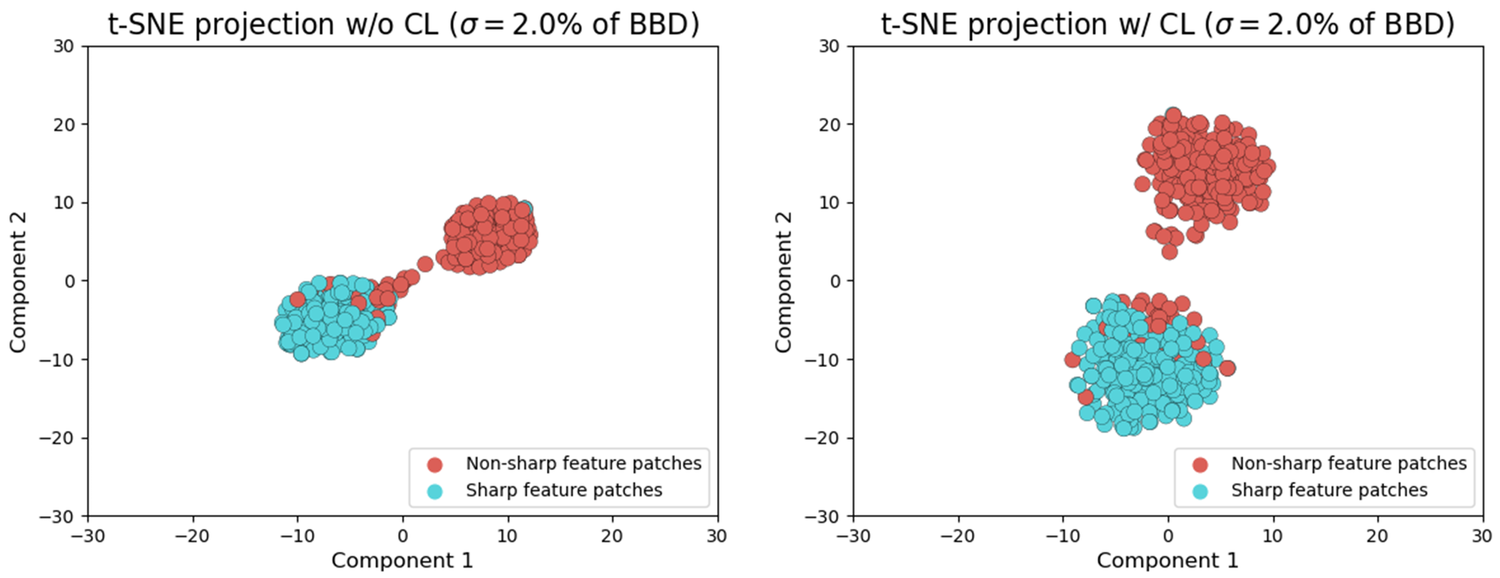}
\caption{T-SNE projections of latent representations of sharp feature patches and non-sharp feature patches. On the left, projections are for latent representations generated by a feature encoder without contrastive learning-based (CL) pretraining and on the right, projections generated by a feature encoder pretrained using contrastive learning. The pretrained feature encoder is able to better distinguish between sharp feature and non-sharp feature patches.}
\label{fig:multi-patch-tsne}
\end{figure}
Contrastive pretraining can be used to better distinguish between latent representations of sharp feature patches from non-sharp feature patches. We can see this in Fig.~\ref{fig:multi-patch-tsne} where the t-SNE projections of latent representations of sharp feature patches are better separated from those of non-sharp feature patches when a feature encoder with contrastive pretraining is used. Here, we looked at 250 sharp feature and 250 non-sharp feature patches, extracted from the Cube shape, and the projections of their respective representations.    

We evaluate a variant without the contrastive learning stage on the validation set to gauge the effect of contrastive learning. This is done by training a regressor with a feature encoder whose weights are randomly initialized and updated along with the weights of the regressor.  
As given by  Table~\ref{tab:ablation-one-task}, the model with contrastive learning achieves a MSAE of 0.034 and a Chamfer distance of 1.431$\times10^{-5}$. The network trained without contrastive learning has a MSAE of 0.037 and a Chamfer distance of 1.445$\times10^{-5}$, indicating the performance gain from contrastive learning. 

\midsepremove
\begin{table}[!ht]
\setlength\tabcolsep{3.5pt}
\centering
\caption{Performance of regression network (1) on the test set when trained on the limited DSP2 data with its feature encoder pretrained on DSP1 using contrastive learning (CL). We compare this to regression network (2) trained on DSP2, without pretraining its feature encoder.}
\label{tab:contrastive-pretraining}
\begin{tabular}{l|c|c}
\toprule
Trained model & MSAE & CD ($\times10^{-5}$) \\
\midrule
(1) Regr. network trained on DSP2 w/ CL & \textbf{0.1660} & \textbf{3.961} \\
(2) Regr. network trained on DSP2 wo/ CL & \uline{0.1713} & \uline{4.057} \\
\midrule
\end{tabular}
\end{table}
\midsepdefault

Next, we look at the effect of utilizing a larger dataset for pretraining the feature encoder while training the regressor on a limited amount of labeled data. To achieve this, we partition our training set into 2 parts. Dataset Partition 1 (DSP1) consists of 18 shapes and is used to train the the feature encoder. Dataset Partition 2 (DSP2) consists of the remaining 4 shapes and is only used to train the regressor. Again, we train a variant from scratch, without the contrastive pretraining, given by (2) in Table~\ref{tab:contrastive-pretraining}. We evaluate the normal estimation and filtering performance on the test set. We see that, with limited training data for the regression task, the positive impact of the contrastive pretraining becomes more apparent.

Finally, we investigate the impact of pretraining the feature encoder on different noise patterns. Therefore, in addition to the Gaussian noise pattern, we augment DSP1 with simulated Lidar noise. In order to achieve this, we utilize the Blensor toolkit~\cite{Blensor} to create noisy variants of the initial 18 shapes of DSP1. The regressor is trained on the limited DSP2 point clouds containing only Gaussian noise, i.e., simulated Lidar noise is not seen by the regressor during training. Thereafter, we look at filtering results on simulated Lidar point clouds with the noise level set to 1.5\%. We compare these results with network (1), with contrastive pretraining on DSP1 with only Gaussian noise and (2), trained from scratch on DSP2. The network trained without contrastive learning, (2) has a Chamfer distance of 3.796$\times10^{-5}$. Network (1), pretrained on DSP1 with noisy point clouds containing only Gaussian noise, fairs marginally better with a Chamfer distance of 3.787$\times10^{-5}$. However, for network (3) pretrained on DSP1 with both Gaussian and simulated Lidar noise patterns, the average Chamfer distance is 3.731$\times10^{-5}$. This indicates the power of pretraining the feature encoder on additional noise patterns.

\subsection{Ablation study on hyperparameters}
\label{sec:ablation-hyperparams}

The three main hyper-parameters which control the predictive power of the network are the following: 
\begin{itemize}
    \item $\alpha$, the weight controlling the relative contributions of the position and normal losses to the overall final loss, Eq~\eqref{eq:joint-loss-function}, during training of the regression network,
    \item $\gamma$, the exponent of the cosine similarity term in Eq.~\eqref{eq:normal-loss} of the main paper, which penalizes the angle difference between predicted and ground-truth normals,
    \item and the contrastive batch size (CBS) used during training of the feature encoder.
\end{itemize} 
We use MSAE as our main evaluation metric for the ablation study as it typically provides a larger degree of agreement with qualitative (visual) comparisons~\cite{Lu-Mesh-Sharp-Feature-Denoising}. Furthermore, we observe a noticeably large deviation in MSAE due to the choice of hyperparameters, as opposed to the deviation in Chamfer distance values.

\begin{table}[!ht]
\setlength\tabcolsep{3.5pt}
\renewcommand{\arraystretch}{1.2}
\centering
\caption{MSAE results on the validation set for different contrastive batch sizes and different values of $\gamma$. Here, $\alpha$ is set to 0.9.}
\midsepremove
\begin{tabular}{l|c|ccc}
\toprule
\multirow{4}{*}{\pbox{20cm}{MSAE}} & \diagbox{\bm{$\gamma$}}{\text{CBS}} & 32 & 128 & 512 \\
\cmidrule{2-5}
& 8  & 0.038 & 0.038 & 0.037 \\
& 10 & 0.036 & 0.036 & 0.036 \\
& 12 & \uline{0.035} & 0.039 & \textbf{0.034} \\
\bottomrule
\end{tabular}
\label{tab:ablation-gamma-cbs}
\end{table}
\midsepdefault

\midsepremove
\begin{table}[!ht]
\renewcommand{\arraystretch}{1.0}
\centering
\caption{MSAE results on the validation set for different values of $\alpha$ with $\gamma$=12 and a contrastive batch size of 512.}
\begin{tabular}{l|ccccc}
\toprule
$\alpha$ & 0.8      & 0.85     & 0.9     & 0.92     & 0.95   \\
\midrule
MSAE & 0.038 & 0.037 & \textbf{0.034} & \uline{0.035} & 0.037 \\
\bottomrule
\end{tabular}
\label{tab:ablation-alpha}
\end{table}
\midsepdefault

In order to find the optimal values of $\gamma$ and the contrastive batch size, we perform a grid search over both. We search over $\gamma$ values of 8, 10 and 12 and CBS values of 32, 128 and 512. Each respective network, for a given pair of values, is evaluated on the validation set and corresponding MSAE values are given in Table~\ref{tab:ablation-gamma-cbs}. Here, it was necessary to set a value of $\alpha$ to perform the grid search. We set $\alpha=0.9$ based on empirical results. The pair (12, 512) of $\gamma$, CBS values optimizes MSAE results. Once the best $\gamma$ and CBS values are determined, we use them to gauge the effect of varying $\alpha$ and justify our selection of $\alpha=0.9$ as demonstrated in Table~\ref{tab:ablation-alpha}. The triplet of values $\alpha=0.9$, $\gamma=12$ and $\text{CBS}=512$ indeed minimizes the MSAE. 

\midsepremove
\begin{table}[!ht]
\setlength\tabcolsep{3.5pt}
\renewcommand{\arraystretch}{1.2}
\centering
\caption{MSAE results on the validation set for different $\beta$ and $\delta$. Here, $\alpha=0.9$, $\gamma=12$ and $\text{CBS}=512$.}
\label{tab:ablation-beta-delta}
\begin{tabular}{l|c|ccc}
\toprule
\multirow{4}{*}{\pbox{20cm}{MSAE}} & \diagbox{~\bm{$\beta$}~}{~\bm{$\delta$}~} & 0.2 & 0.3 & 0.4 \\
\cmidrule{2-5}
& 0.01 & 0.036 & \textbf{0.034} & \uline{0.035} \\
& 0.02 & \uline{0.035} & \textbf{0.034} & 0.036 \\
& 0.03 & \uline{0.035} & 0.037 & 0.036 \\
\bottomrule
\end{tabular}
\end{table}
\midsepdefault

Next we perform additional ablation studies to confirm our chosen values of $\beta$ and $\delta$ which appear in Eq.~\eqref{eq:position-loss} and Eq.~\eqref{eq:normal-loss}, respectively. Table~\ref{tab:ablation-beta-delta}  displays the validation results. The values $\beta=0.01$ and $\delta=0.3$ minimize the MSAE.

\subsection{Contrastive learning without rotational augmentation}
\label{sec:contrastive-learning-no-rotation}
We consider the case of a feature encoder trained without applying the second augmentation to contrastive patches $\mathcal{Q}$. Therefore, the only augmentation applied to generate augmented pairs is noise corruption. For such an encoder, we retrain the regression network with hyperparameters $\alpha=0.9$, $\beta=0.01$, $\delta=0.3$, $\gamma=12$ and $\text{CBS}=512$. We recover a MSAE of 0.037, compared to a lower MSAE of 0.034 with the second, rotational, augmentation. This indicates the importance of the rotation augmentation to the representation learning process.

\subsection{Evaluation without post-processing refinement} 
\label{sec:no-post-processing}
Finally, we observed our iterative filtering strategy would shrink the filtered point cloud, which necessitates Taubin smoothing-like inflation (TS). PointCleanNet (PCN)~\cite{Rakotosaona-PCN} also used a similar post-step, so we compare with it. Furthermore, to maximize the interaction between normal estimation and point cloud filtering, a point update strategy like that in~\cite{Lu-Low-Rank} (LRMA) is used. Previous research rarely exploits this interconnected relationship between normal estimation and filtering. Table~\ref{tab:ablation-post-proc} shows that Taubin smoothing is important to PCN and our method, and our pure network output is better than PCN's pure network output (CD: 2.10$\times10^{-5}$ vs 2.83$\times10^{-5}$).
\midsepremove
\begin{table}[!tp]
\centering
\caption{Results on the validation set for PCN and our method when using post-processing Taubin Smoothing (TS) and Low Rank Matrix Approximation (LRMA) position update. Our method outperforms PCN on the filtering task, with and without the Taubin smoothing-like post processing step. Moreover, our method, unlike PCN, estimates point normals that are then used to further improve filtering results.}
\begin{tabular}{l|l|cc}
\toprule
& Post Processing & PCN & Ours \\
\midrule
\multirow{3}{*}{\pbox{100pt}{MSAE}} & None & N/A & \uline{0.054} \\
& TS & N/A & \textbf{0.034} \\
& TS+LRMA & N/A & \textbf{0.034} \\
\midrule
\multirow{3}{*}{\pbox{100pt}{CD \\($\times10^{-5}$)}} & None & 2.83 & 2.10 \\
& TS & 1.71 & \uline{1.63} \\
& TS+LRMA & N/A & \textbf{1.43} \\
\bottomrule
\end{tabular}
\label{tab:ablation-post-proc}
\end{table}
\midsepdefault

\section{Limitations} 
\begin{figure}[!h]
\centering
\includegraphics[width=2.4in]{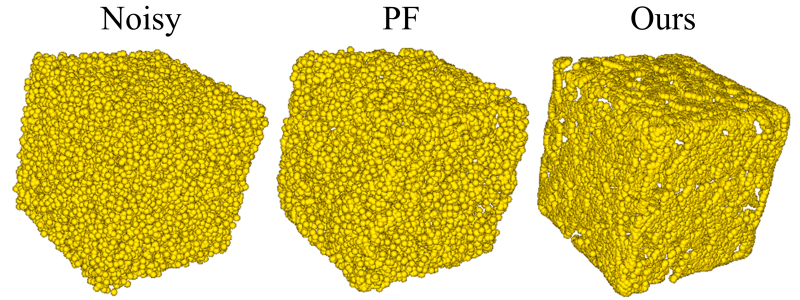}
\caption{Visual results for the Cube shape with 10K points and 0.8\% Gaussian noise.  }
\label{fig:filtering-limitation}
\end{figure}
Similar to previous deep learning methods~\cite{Zhang-Pointfilter}, our method cannot handle sparsely sampled point clouds well. Fig.~\ref{fig:filtering-limitation}  demonstrates the filtering results for the Cube shape with 10K points (0.8\% Gaussian noise). Our method is able to recover some sharp features such as edges, yet fails to recover the planar surfaces. Pointfilter, by comparison, does not recover edge information.

\section{Conclusion}

In this paper, we introduced a deep learning method that jointly learns point displacements and their respective normals from point cloud data. This approach of a simultaneous inference of both position and normal information yields accurate filtered positions and normals. Our method displays an ability to preserve sharp features and fine details. This ability is derived from the 3D patch based contrastive learning 
which faithfully outputs patch representations based on their distinct geometric structure and the introduced joint loss that encourages the effective learning of normal and position information. We conduct extensive experiments and show that our method generally performs better than state-of-the-art methods in terms of both point cloud filtering and normal estimation. It also generalizes well between both CAD-like and non-CAD-like point clouds. 



\ifCLASSOPTIONcaptionsoff
  \newpage
\fi

\bibliographystyle{IEEEtran}
\bibliography{IEEEabrv,egbib}
\vskip -2\baselineskip plus -1fil
\begin{IEEEbiography}[{\includegraphics[width=1in,height=1.25in,clip,keepaspectratio]{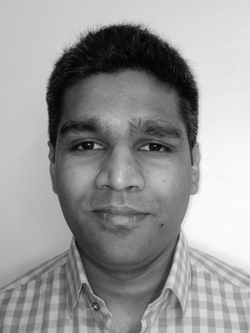}}]{Dasith de Silva Edirimuni}
received the BSc degree in Physics from Hardin-Simmons University, Texas, USA in 2013 and the MSc degree in Physics from the University of Melbourne, Australia, in 2017. In 2020 he received the Master of Information Technology degree from Swinburne University, Australia and is currently a PhD student in Information Technology at Deakin University, Australia. His research interests include 3D computer vision areas such as point cloud filtering and scene understanding.
\end{IEEEbiography}
\vskip -2\baselineskip plus -1fil
\begin{IEEEbiography}[{\includegraphics[width=1in,height=1.25in,clip,keepaspectratio]{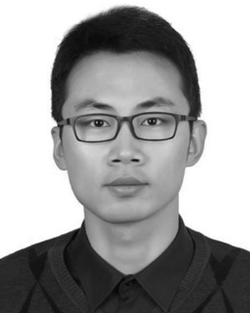}}]{Xuequan Lu}
received the PhD degree from Zhejiang University, China, in June 2016. He is currently an assistant professor with the School of Information Technology, Deakin University, Australia. He has spent more than two years as a research fellow in Singapore. His research interests include visual
computing, for example, geometry modeling, processing and analysis, animation or simulation, and 2D data processing and analysis. He was a member of the International Program Committee of GMP 2021, a PC member in CVM 2020, and a TPC member in ICONIP 2019.
\end{IEEEbiography}
\vskip -2\baselineskip plus -1fil
\begin{IEEEbiography}[{\includegraphics[width=1in,height=1.25in,clip,keepaspectratio]{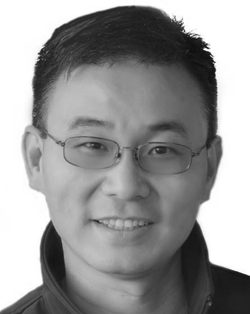}}]{Gang Li}
(Senior Member, IEEE) is currently an
Associate Professor with the School of Information
Technology, Deakin University, Melbourne, VIC,
Australia. His research interests include data mining, data privacy, causal discovery, and business
intelligence.
\end{IEEEbiography}
\vskip -2\baselineskip plus -1fil
\begin{IEEEbiography}[{\includegraphics[width=1in,height=1.25in,clip,keepaspectratio]{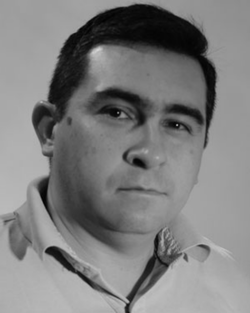}}]{Antonio Robles-Kelly}
received the BEng degree in Electronics and Telecommunications with honours in 1998 and the PhD degree in Computer Science from the University of York, United Kingdom, in 2003. He was in York until December 2004 as a research associate under the MathFitEPSRC framework. In 2005, he moved to Australia and took a research scientist appointment with National ICT Australia (NICTA). In 2006 he became the project leader of the Imaging Spectroscopy team at NICTA and, from 2007 to 2009, he was a postdoctoral research fellow of the Australian Research Council. In 2016, he joined CSIRO where he is a principal researcher with Data61 and, in 2018, became a Machine Learning and Artificial Intelligence professor at Deakin University, Australia. His research has been applied to areas such as biosecurity, forensics, food quality assurance, and biometrics and is now being deployed by CSIRO under the trademark of Scyllarus (www.scyllarus.com). He has served as the president of the Australian Pattern Recognition Society (APRS) and is an associate editor of the Pattern Recognition Journal and IET Computer Vision Journal. He is a senior member of the IEEE, the president of the TC2 (Technical Committee on structural and syntactical pattern recognition) of the International Association for Pattern Recognition (IAPR) and an adjunct associate professor at the ANU. He has also been a technical committee member, area, and general chair of several mainstream computer vision and pattern recognition conferences.
\end{IEEEbiography}




\setcounter{section}{0}
\renewcommand\thesection{\Alph{section}}
\renewcommand\thesubsection{\thesection.\arabic{subsection}}

\title{Contrastive Learning for Joint Normal Estimation\\and Point Cloud Filtering: Supplementary}

\author{Dasith~de~Silva~Edirimuni,~
        Xuequan~Lu,~\IEEEmembership{Senior~Member,~IEEE,}
        Gang~Li,~\IEEEmembership{Senior~Member,~IEEE,}
        and~Antonio~Robles-Kelly,~\IEEEmembership{Senior~Member,~IEEE}
\IEEEcompsocitemizethanks{\IEEEcompsocthanksitem D. de Silva Edirimuni, X. Lu, G. Li and A. Robles-Kelly are with the School of Information Technology, Deakin University, Waurn Ponds, Victoria, 3216, Australia (e-mail: \{dtdesilva, xuequan.lu, gang.li, antonio.robles-kelly\}@deakin.edu.au). A. Robles-Kelly is also with the Defense Science and Technology Group, Australia.}
\thanks{Manuscript received Month Day, Year; revised Month Day, Year.}
\thanks{(Corresponding author: Xuequan Lu.)}}

\markboth{Journal of \LaTeX\ Class Files,~Vol.~X, No.~X, Month~Year}%
{de Silva Edirimuni \MakeLowercase{\textit{et al.}}: Contrastive Learning for Joint Normal Estimation and Point Cloud Filtering: Supplementary}

\maketitle

\IEEEdisplaynontitleabstractindextext

%
\IEEEpeerreviewmaketitle

Here we provide supplementary material to the main paper. In particular, we provide additional information on the following:
\begin{enumerate}
\item Sharp feature points classification
\item Varying density point clouds
\item Further comparisons on synthetic and real-world scan data
\end{enumerate}

\section{Sharp Feature Points Classification}
\label{app:sharp-features}
\begin{figure}[!th]
\centering
\includegraphics[width=0.4\textwidth]{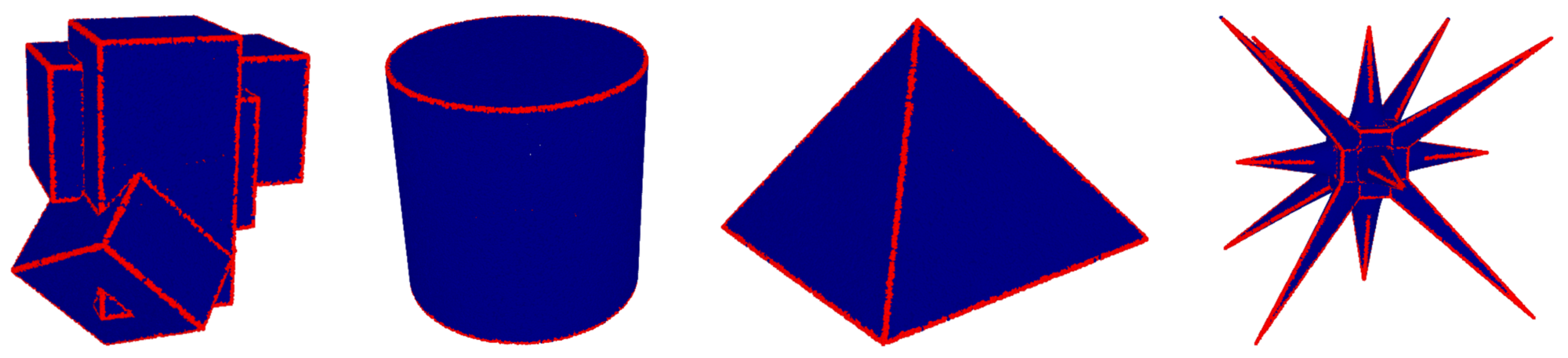}
\caption{Sharp feature points are given in red.}
\label{fig:sharp-features}
\end{figure}
In order to explicitly analyze the performance of normal estimation and filtering methods at sharp features on our synthetic data, we must first determine which points correspond to sharp features, i.e., points that lie on anisotropic surfaces. To achieve this, we exploit the ground truth normal information for each clean synthetic point cloud. For each point in the clean point cloud, we consider neighborhoods of their 10 nearest neighbors. Nearest neighbors whose normals make an angle $\pi/6<\theta<5\pi/6$ with the neighborhood's central point's normal are classified as feature points. Here, the threshold angle between normals for determining a sharp feature point is $\pi/6$. If the angle is below the threshold, the surface varies smoothly. An angle greater than $5\pi/6$ most likely corresponds to a normal of a point on a surface parallel to that of the central point and is not considered. This scheme allows us to extract points along sharp edges and corners. The unique indices corresponding to these points (on ground truth and filtered point clouds) are used to determine MSAE and Point2Surface distance values at sharp features.

\section{Varying density point clouds}
\begin{figure}[!th]
\centering
\includegraphics[width=0.4\textwidth]{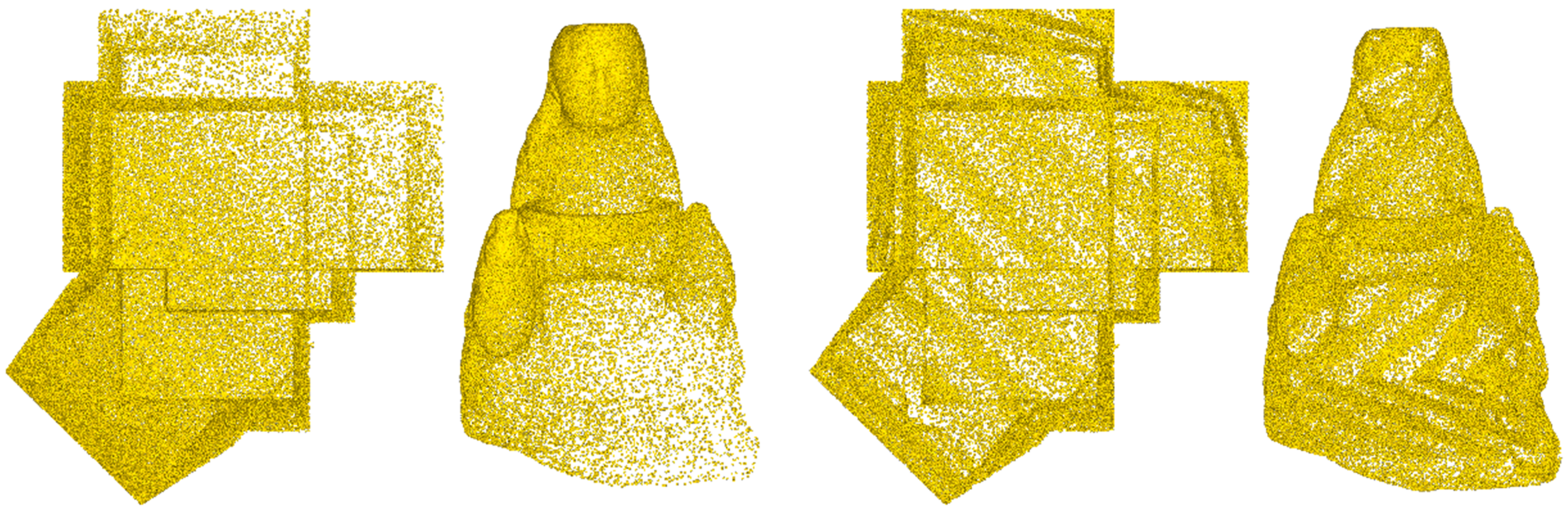}
\caption{Left: Point clouds sampled with the gradient varying density regime. Right: Point clouds sampled with the striped varying density regime.}
\label{fig:varying-density}
\end{figure}
Varying density point clouds are obtained by applying two different sampling regimes, gradient and striped, to the original test meshes. Fig.~\ref{fig:varying-density} illustrates point clouds from these two regimes. Thereafter, Gaussian noise of $\sigma=0.8\%$ is added to each point cloud in order to evaluate the robustness of normal estimation and filtering methods on noisy, varying density point clouds.

\section{Further comparisons on synthetic and real-world scan data}
\begin{figure*}[!tp]
\centering
\includegraphics[height=0.9\textheight]{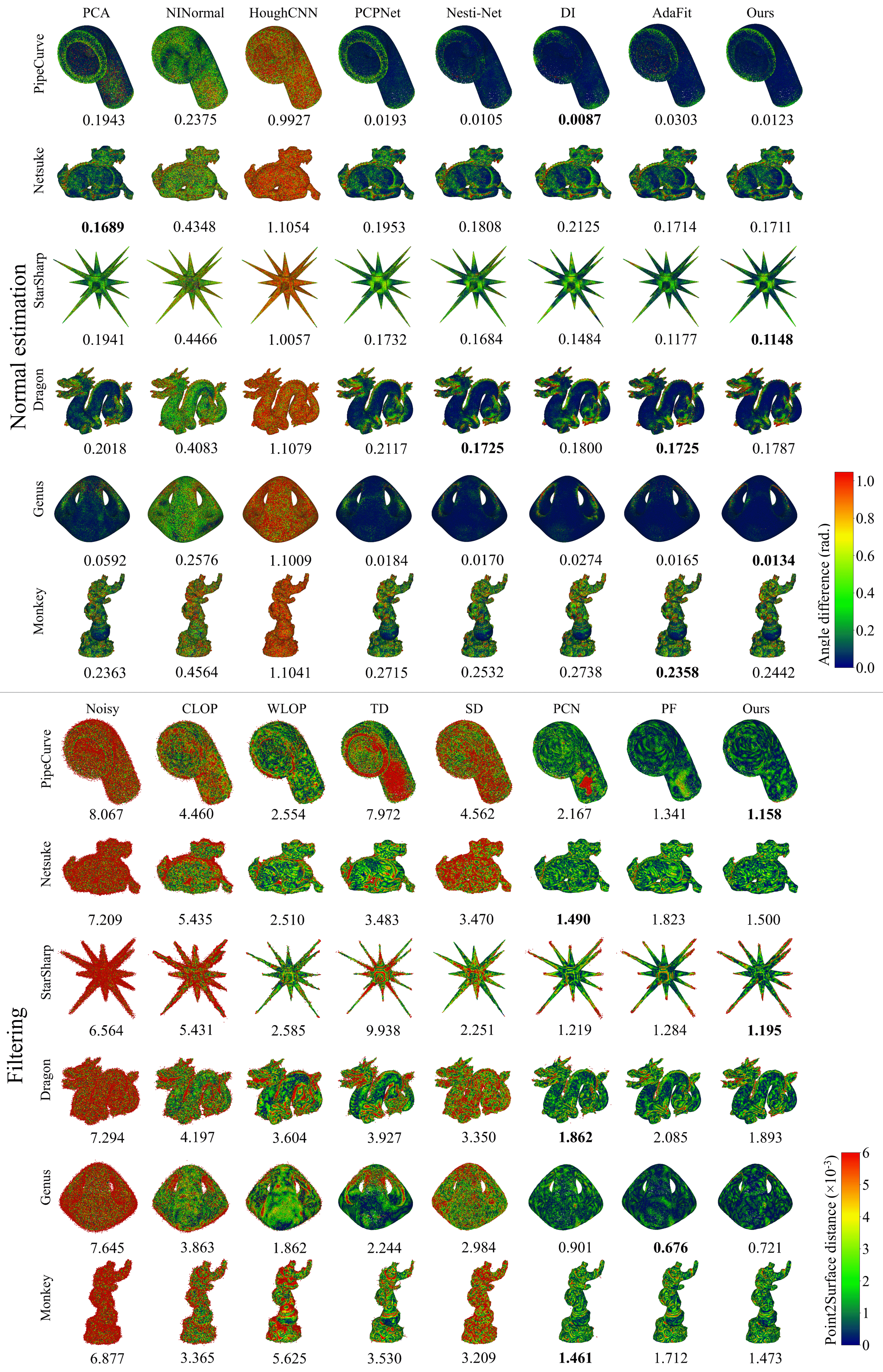}
\caption{Normal estimation (top half) and filtering (bottom half) results on shapes with 0.8\% Gaussian noise with respect to the bounding box diagonal. For normal estimation, the respective mean squared angular error (MSAE) is given below each shape and the heat map corresponds to the angle difference at each point. For filtering, the Chamfer distance ($\times 10^{-5}$) is given below each shape and the heat map  corresponds to the scale normalized Point2Surface distance ($\times 10^{-3}$).}
\label{fig:all-synthetic-results-additional}
\end{figure*}

Fig.~\ref{fig:all-synthetic-results-additional} displays normal estimation and filtering results on additional shapes within our synthetic data for Gaussian noise with standard deviation of $0.8\%$ of the bounding box diagonal and Table~\ref{tab:additional-results} details performance across all noise levels, for each respective shape. In the normal estimation task, we outperform other methods and generalize well between CAD shapes such as StarSharp and non-CAD shapes such as Netsuke, especially at higher noise levels. For the filtering task, we perform competitively, and our method is able to reliably recover sharp features on CAD shapes such as StarSharp and PipeCurve and fine details on non-CAD shapes such as Netsuke.

\midsepremove
\begin{table*}[!tp]
\centering
\caption{Individual MSAE and Chamfer distance values for shapes presented in Fig.~\ref{fig:all-synthetic-results-additional}, at different noise levels. The standard deviation $\sigma$ is with respect to the bounding box diagonal of the clean point cloud. Top results in bold and second best results are underlined.}
\begin{tabular}{c|c|ccccc|cccc}
\toprule
\multirow{2}{*}{Shape} & \multicolumn{1}{c|}{\multirow{2}{*}{$\sigma$}} & \multicolumn{5}{c|}{Normal estimation - MSAE} & \multicolumn{4}{c}{Filtering - Chamfer distance ($\times10^{-5}$)} \\
\cmidrule{3-11}
 & \multicolumn{1}{c|}{} & PCPNet & Nesti-Net & DI & AdaFit & Ours & Noisy & PCN & PF & Ours \\
 \midrule
\multirow{5}{*}{PipeCurve} & 0.6\% & 0.0169 & \uline{0.0068} & \textbf{0.0056} & 0.0146 & 0.0115 & 5.190 & 1.482 & \uline{1.015} & \textbf{1.024} \\
 & 0.8\% & 0.0193 & \uline{0.0105} & \textbf{0.0087} & 0.0303 & 0.0123 & 8.067 & 2.167 & \uline{1.341} & \textbf{1.158} \\
 & 1.1\% & 0.0230 & 0.0173 & \uline{0.0132} & 0.0689 & \textbf{0.0121} & 12.922 & 5.041 & \uline{4.691} & \textbf{1.651} \\
 & 1.5\% & 0.0304 & 0.0324 & \uline{0.0214} & 0.0597 & \textbf{0.0196} & 19.805 & 11.395 & \uline{11.26} & \textbf{3.214} \\
 & 2.0\% & 0.0398 & 0.0433 & \uline{0.0280} & 0.0421 & \textbf{0.0272} & 30.269 & 25.024 & \uline{16.563} & \textbf{7.103} \\
 \midrule
\multirow{5}{*}{Netsuke} & 0.6\% & 0.1628 & \uline{0.1552} & 0.1793 & \textbf{0.1440} & 0.1591 & 4.555 & \uline{1.263} & 1.545 & \textbf{1.237} \\
 & 0.8\% & 0.1953 & 0.1808 & 0.2125 & \uline{0.1714} & \textbf{0.1712} & 7.209 & \uline{1.49} & 1.823 & \textbf{1.500} \\
 & 1.1\% & 0.2462 & 0.2163 & 0.2526 & \uline{0.2155} & \textbf{0.1964} & 12.209 & \uline{2.160} & 2.294 & \textbf{2.040} \\
 & 1.5\% & 0.3122 & \uline{0.2736} & 0.3054 & 0.2763 & \textbf{0.2459} & 20.832 & 4.585 & \textbf{3.171} & \uline{3.826} \\
 & 2.0\% & 0.3714 & \uline{0.3447} & 0.3631 & 0.3463 & \textbf{0.3044} & 33.812 & 14.521 & \textbf{5.077} & \uline{9.593} \\
 \midrule
\multirow{5}{*}{StarSharp} & 0.6\% & 0.1371 & 0.1257 & \uline{0.1029} & \textbf{0.0801} & 0.1039 & 4.002 & 0.93 & \uline{0.895} & \textbf{0.822} \\
 & 0.8\% & 0.1732 & 0.1684 & 0.1484 & \uline{0.1177} & \textbf{0.1148} & 6.564 & \uline{1.219} & 1.284 & \textbf{1.195} \\
 & 1.1\% & 0.2280 & 0.2309 & 0.1977 & \uline{0.1671} & \textbf{0.1338} & 11.572 & 2.416 & \textbf{1.980} & \uline{2.597} \\
 & 1.5\% & 0.2776 & 0.2811 & 0.2558 & \uline{0.2226} & \textbf{0.1622} & 20.749 & 7.205 & \textbf{3.535} & \uline{6.190} \\
 & 2.0\% & 0.3219 & 0.3206 & 0.3020 & \uline{0.2849} & \textbf{0.1696} & 36.616 & 20.958 & \textbf{5.770} & \uline{14.617} \\
 \midrule
\multirow{5}{*}{Dragon} & 0.6\% & 0.1562 & \textbf{0.1302} & 0.1337 & \uline{0.1310} & 0.1526 & 4.644 & \uline{1.494} & 1.613 & \textbf{1.443} \\
 & 0.8\% & 0.2117 & \uline{0.1725} & 0.1800 & \textbf{0.1725} & 0.1787 & 7.294 & \uline{1.862} & 2.085 & \textbf{1.893} \\
 & 1.1\% & 0.2857 & \uline{0.2279} & 0.2329 & 0.2287 & \textbf{0.2087} & 12.183 & \uline{2.858} & 2.863 & \textbf{2.609} \\
 & 1.5\% & 0.3765 & \uline{0.3170} & 0.3301 & 0.3245 & \textbf{0.2772} & 20.215 & 6.577 & \uline{4.491} & \textbf{4.394} \\
 & 2.0\% & 0.4768 & 0.4279 & 0.4386 & \uline{0.4240} & \textbf{0.3684} & 32.734 & 19.231 & \textbf{6.269} & \uline{9.000} \\
 \midrule
\multirow{5}{*}{Genus} & 0.6\% & 0.0128 & 0.0119 & 0.0218 & \textbf{0.0103} & \uline{0.0112} & 4.720 & 0.817 & \uline{0.642} & \textbf{0.638} \\
 & 0.8\% & 0.0184 & \uline{0.0170} & 0.0274 & 0.0165 & \textbf{0.0134} & 7.645 & 0.901 & \textbf{0.676} & \uline{0.721} \\
 & 1.1\% & 0.0318 & \uline{0.0303} & 0.0394 & 0.0370 & \textbf{0.0178} & 13.249 & 1.370 & \textbf{0.909} & \uline{0.973} \\
 & 1.5\% & 0.0570 & \uline{0.0486} & 0.0526 & 0.0628 & \textbf{0.0248} & 22.718 & 3.647 & \uline{2.025} & \textbf{2.000} \\
 & 2.0\% & 0.1022 & 0.0865 & \uline{0.0724} & 0.1119 & \textbf{0.0478} & 37.945 & 14.607 & \textbf{3.835} & \uline{6.064} \\
 \midrule
\multirow{5}{*}{Monkey} & 0.6\% & 0.2225 & \uline{0.2113} & 0.2298 & \textbf{0.1953} & 0.2174 & 4.371 & \uline{1.220} & 1.393 & \textbf{1.150} \\
 & 0.8\% & 0.2715 & 0.2532 & 0.2738 & \textbf{0.2358} & \uline{0.2442} & 6.877 & \uline{1.461} & 1.712 & \textbf{1.473} \\
 & 1.1\% & 0.3372 & 0.3083 & 0.3297 & \uline{0.2964} & \textbf{0.2788} & 11.706 & \uline{2.112} & 2.175 & \textbf{1.978} \\
 & 1.5\% & 0.4125 & \uline{0.3719} & 0.3948 & 0.3745 & \textbf{0.3326} & 19.968 & 4.956 & \textbf{2.896} & \uline{3.621} \\
 & 2.0\% & 0.4856 & \uline{0.4516} & 0.4754 & 0.4626 & \textbf{0.4103} & 32.578 & 16.503 & \textbf{4.651} & \uline{9.033} \\
 \bottomrule
\end{tabular}
\label{tab:additional-results}
\end{table*}
\midsepdefault

\begin{figure*}[!tp]
\centering
\includegraphics[width=0.6\textwidth]{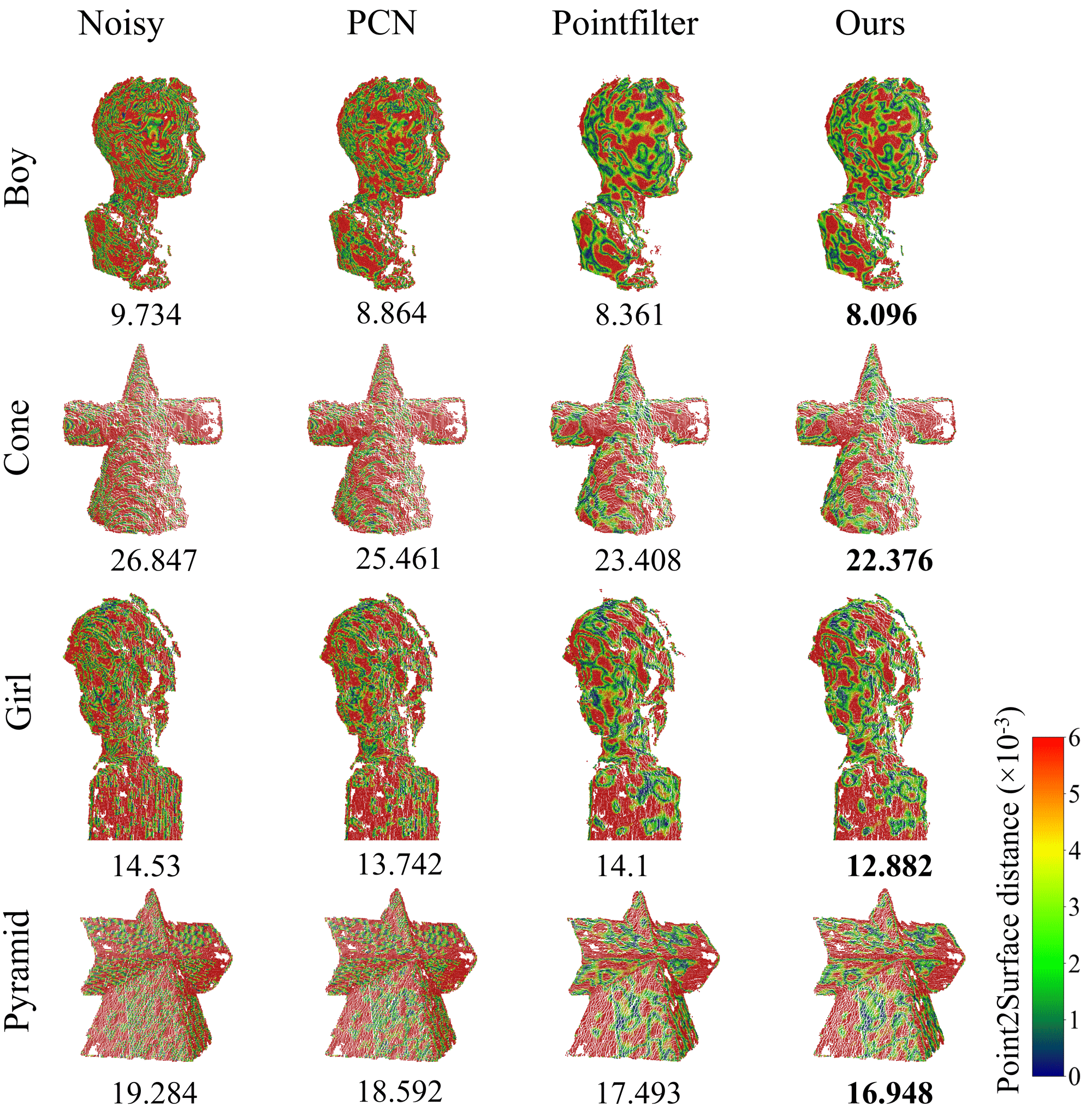}
\caption{Filtering results on the Kinect v1 and v2 datasets, the Chamfer distance ($\times 10^{-5}$) is given below each shape and the heat map  corresponds to the scale normalized Point2Surface distance ($\times 10^{-3}$).}
\label{fig:kinect}
\end{figure*}

\begin{figure*}[!tp]
\centering
\includegraphics[width=\textwidth]{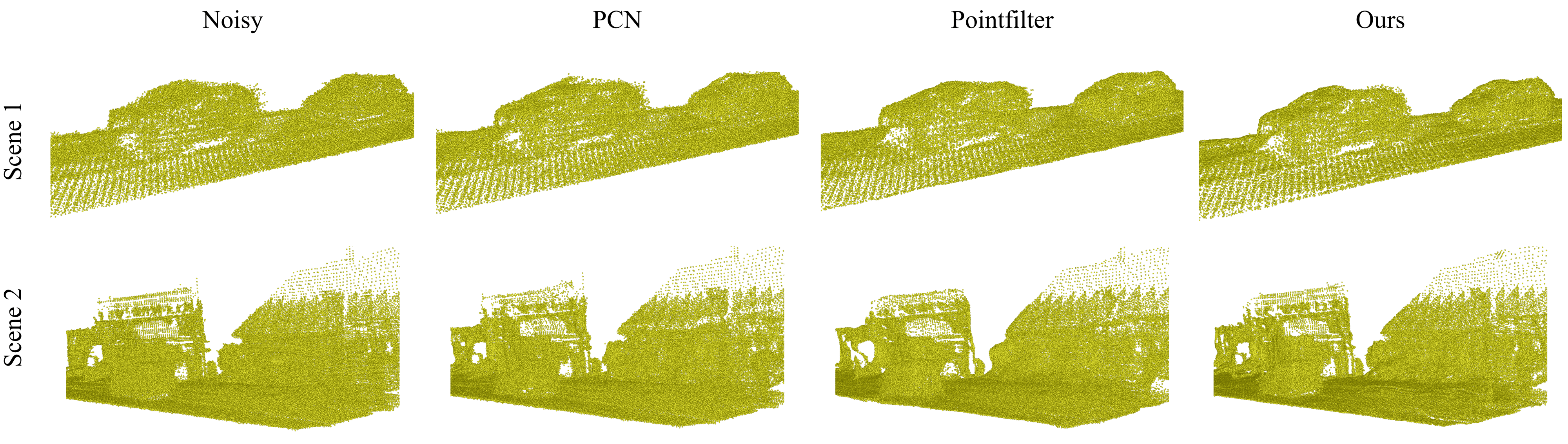}
\caption{Visual filtering results on 2 scans from the Kitti-360 dataset. Scene 1 corresponds to vehicles parked on grass, next to a road while scene 2 depicts parked vehicles and houses along a street.}
\label{fig:kitti-360}
\end{figure*}

On the Kinect v1 and v2 datasets, we outperform other methods as depicted by Fig.~\ref{fig:kinect}. The average Chamfer distance and Point2Surface results are given in Table 4 of the main paper. Fig.~\ref{fig:kitti-360} provides an evaluation on 2 scenes of the Kitti-360 dataset. This dataset contains sparse point clouds with high amounts of real world noise which makes it difficult to filter these scenes. However, as shown in scene 1, we perform better at filtering noise and retrieving the underlying shapes of parked cars as compared to other methods.

\end{document}